\documentclass{article}

\PassOptionsToPackage{numbers}{natbib}

\usepackage[preprint]{neurips_2023}

\usepackage[utf8]{inputenc} 
\usepackage[T1]{fontenc}    
\usepackage{url}            
\usepackage{booktabs}       
\usepackage{amsfonts}       
\usepackage{nicefrac}       
\usepackage{microtype}      
\usepackage[usenames,dvipsnames]{xcolor}
\usepackage[colorlinks,linktoc=all]{hyperref}
\hypersetup{citecolor=Blue}
\hypersetup{linkcolor=Blue}
\hypersetup{urlcolor=Blue}

\usepackage{microtype}
\usepackage{graphicx}
\usepackage{subcaption}
\usepackage{mfirstuc}
\usepackage{booktabs} 
\usepackage{color}
\usepackage{natbib}

\usepackage{hyperref}

\usepackage{amsmath}
\usepackage{amssymb}
\usepackage{mathtools}
\usepackage{amsthm}
\usepackage{multirow}
\usepackage{listings}
\usepackage{menukeys} 

\usepackage[capitalize,noabbrev]{cleveref}

\theoremstyle{plain}
\newtheorem{theorem}{Theorem}[section]

\newtheorem{lemma}[theorem]{Lemma}

\theoremstyle{definition}

\theoremstyle{remark}

\newcommand\pl[1]{}
\newcommand\qian[1]{}
\newcommand\sarah[1]{}
\newcommand\eric[1]{}

\renewcommand\pl[1]{\textcolor{purple}{[PL: #1]}}
\renewcommand\qian[1]{\textcolor{blue}{[Qian: #1]}}
\renewcommand\sarah[1]{\textcolor{orange}{[SC: #1]}}
\renewcommand\eric[1]{\textcolor{red}{[EZ: #1]}}

\newcommand{\modelname}{lexinvariant\space}

\newcommand{\Modelname}{Lexinvariant\space}

\title{Lexinvariant Language Models}

\author{%
  Qian Huang$^{1}$ \\
  \texttt{qhwang@cs.stanford.edu} \\
  \And
   Eric Zelikman$^{1}$ \\
  \texttt{ezelikman@cs.stanford.edu} \\
  \And
  Sarah Li Chen$^{1}$ \\
  \texttt{sachen@stanford.edu} \\
  \AND
  Yuhuai Wu$^{12}$ \\
  \texttt{yuhuai@cs.stanford.edu} \\
\And
Gregory Valiant$^{1}$ \\
\texttt{gvaliant@cs.stanford.edu} \\
\And
Percy Liang$^{1}$ \\
\texttt{pliang@cs.stanford.edu} \\
\AND \normalfont{$^1$Stanford University} \\
\normalfont{$^2$Google Research} \\
}

\usepackage[subtle]{savetrees}

\begin{document}

\maketitle

\begin{abstract}
Token embeddings, a mapping from discrete lexical symbols to continuous vectors, are at the heart of any language model (LM). However, lexical symbol meanings can also be determined and even redefined by their structural role in a long context. 
In this paper, we ask: is it possible for a language model to be performant without \emph{any} fixed token embeddings?
Such a language model would have to rely entirely on the co-occurence and repetition of tokens in the context rather than the \textit{a priori} identity of any token. To answer this, we study \textit{\modelname}language models that are invariant to lexical symbols and therefore do not need fixed token embeddings in practice. First, we prove that we can construct a \modelname LM to converge to the true language model at a uniform rate that is polynomial in terms of the context length, with a constant factor that is sublinear in the vocabulary size. Second, to build  a \modelname LM, we simply encode tokens using random Gaussian vectors, such that each token maps to the same representation within each sequence but different representations across sequences. Empirically, we demonstrate that it can indeed attain perplexity comparable to that of a standard language model, given a sufficiently long context. We further explore two properties of the \modelname language models: First, given text generated from a substitution cipher of English, it implicitly implements Bayesian in-context deciphering and infers the mapping to the underlying real tokens with high accuracy. Second, it has on average 4X better accuracy over synthetic in-context reasoning tasks. Finally, we discuss regularizing standard language models towards lexinvariance and potential practical applications. 
\end{abstract}

\section{Introduction}

All language processing systems rely on a stable lexicon, which assumes that a token (a word or subword such as \emph{tree}) has a consistent contribution to the meaning of a text (though of course this meaning is mediated by context).
In neural language models (LMs),
this contribution is the token embedding, which \emph{stably} maps each token into a continuous vector \cite{schutze-1993-part, Mikolov2013EfficientEO, Mikolov2010RecurrentNN, Devlin2019BERTPO, Brown2020LanguageMA}.
However, in real language, a token's contribution might be determined by its structural role; in math and code, novel variable names are arbitrarily defined to carry new meaning, and poems such as Jabberwocky exploit humans' lexical flexibility in interpreting novel words such as \emph{vorpal}. Besides standard language understanding, this lexical flexibility also correlates with a stronger in-context reasoning performance. For example, GPT-3 \cite{Brown2020LanguageMA} and other large language models that demonstrate high lexical flexibility show strong performance on tasks involving in-context reasoning over new concepts and rules. 

\begin{figure}[t]
    \vspace{-1.5em}
    \centering
    \begin{minipage}[h]{0.25\linewidth}
    \centering
    \includegraphics[width=\linewidth]{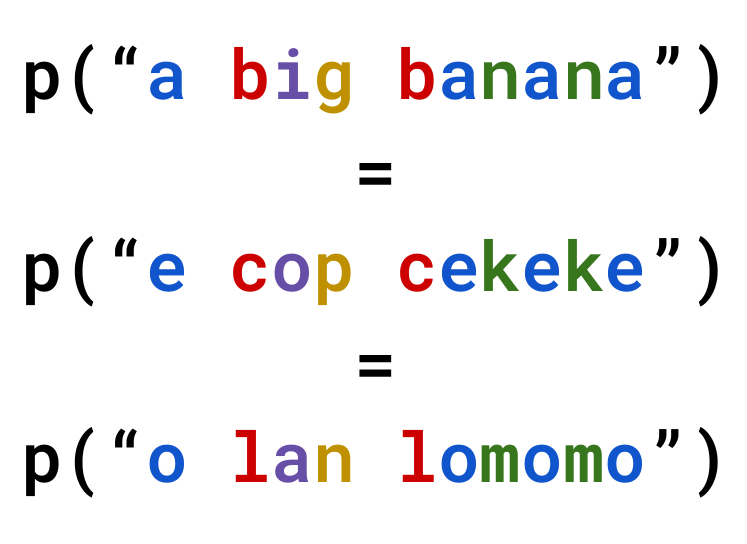}
    \subcaption{Lexinvariance \label{fig:figure1a}}
    \end{minipage}
    \hspace{0.1\linewidth}
    \begin{minipage}[h]{0.3\linewidth}
    \centering
    \includegraphics[width=0.9\linewidth]{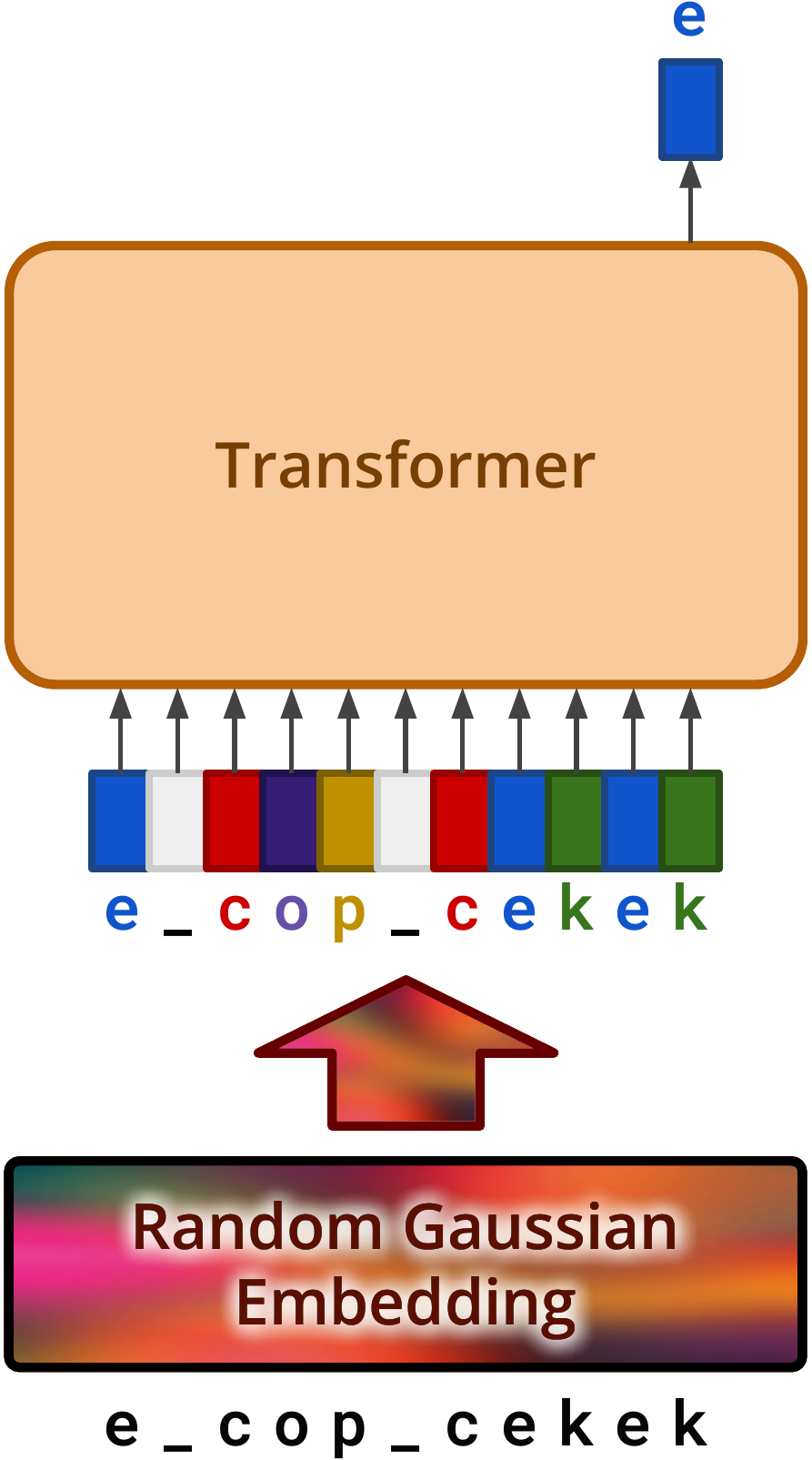}
    \vspace{-0.5em}
    \subcaption{Lexinvariant Language Model \label{fig:figure1b}}
    \end{minipage}
    \vspace{-5px}
    \caption{Definition (a)  and construction (b)  of lexinvariant language model \label{fig:figure1}}
    \vspace{-1em}
\end{figure}

Motivated by the above, we ask whether we can push this flexibility to the extreme: can we build a language model without \emph{any} stable lexical mapping?
To this end, we formulate and study such \emph{\modelname} language models. We define a \modelname language model as a language model that assigns the same probability to all lexical permutations 
of a sequence. Formally, we define a lexical permutation $\pi$ to be a one-to-one mapping of a set of lexical symbols \footnote{We specifically consider lexical symbols as tokens, not necessarily words or other linguistic units.} onto itself. 
Then the \modelname language model is defined as a language model over the symbol sequence $x_1,\dots, x_n$ with the following property: 
\begin{equation}
    p(x_1,\dots, x_n) = p(\pi(x_1),\dots, \pi(x_n)) \;\; \forall \pi \label{definition}
\end{equation}
 For example, a \modelname language model (whose vocabulary is letters and space) should assign the same probability to the phrase ``\texttt{a big banana}'' as ``\texttt{e cop cekeke}'' 
 because the two are the same up to the permutation $\pi = \{ \texttt{a} : \texttt{e}, \texttt{b}: \texttt{c}, \texttt{i}: \texttt{o}, \texttt{n}: \texttt{k}, \texttt{g}: \texttt{p}, \cdots  \}$
(Figure \ref{fig:figure1a}). 

The central question is: how well can lexinvariant language models predict the next token given an increasingly long context?  We find the answer is almost as well as standard language models, both theoretically and empirically.  This is rather surprising given that lexinvariance seems like a strong limitation (a model doesn\'t know what any individual symbol means!)  However, the intuition is that given longer contexts,
a lexinvariant model can both infer the latent permutation $\pi$ (lazily) up to whatever ambiguity is present in the language model, and do the standard next word prediction task jointly.

Theoretically, we prove that a constructed \modelname language model  can  converge to the true language model  as the context length increases---that is, the average L1 distance between the predictions of the two models decreases with a   convergence rate of $O\left( (\frac{d}{T}) ^{\frac{1}{4}}\right)$, where $T$ is the length of the context and $d$ is the vocabulary size, and where the big-O notation hides polylogarithmic factors of $d$ and $T$ and an absolute constant that is \emph{independent} of the language model.

Empirically, we train a \modelname LM by replacing standard embeddings in a decoder-only Transformer \cite{NIPS2017_3f5ee243} with per-sequence random Gaussian vectors, such that the same symbols get the same embedding within each sequence but get different embedding across sequences (Figure~\ref{fig:figure1b}).  
We indeed see that the perplexity gap between the \modelname LM and the standard LM shrinks as context length increases, as shown in Section \ref{sec:convergence}. With a 150M parameters Transformer and a small character-level vocabulary (130 tokens), the average perplexity gap shrinks from 9X to less than 1X the average perplexity of a standard LM after observing 512 tokens over The Pile \cite{Gao2020ThePA}. With a larger 32K vocabulary, the gap also shrinks, especially on the more structured text like GitHub code, albeit at a much slower rate. 

We then explore two additional properties of the \modelname LM: in-context deciphering and symbol manipulation.
First, we show that given a ciphertext generated by applying a substitution cipher to English text, the \modelname LM can be seen as implicitly approximating Bayesian inference of the lexical permutation, i.e., cipher key, in-context. 
To show this empirically, we train a small MLP probe on top of a frozen pretrained \modelname LM to predict the deciphered token corresponding to the last seen cipher token. We can then read out the inferred cipher key with each prefix of the sequence. 
We show that the accuracy of this inferred cipher key  quickly improves as context length grows, reaching 99.6\% average accuracy. We also show examples in Section \ref{sec:decipher} that visualize the uncertainties over different possible lexical mappings maintained by the \modelname LM when the cipher key is ambiguous and that the semantic meaning of a symbol with very rare occurrence can be inferred efficiently relative to other common symbols in context.
Second, we show that \modelname models perform better than traditional models over synthetic pure in-context reasoning tasks that involve symbol manipulation.  We observe a significant 4X improvement over a standard language model.

While the primary motivation of this paper is scientific exploration of a new idea, lexinvariance, we were also curious to see if it could help improve certain tasks, generalizing the performance gain we see on synthetic tasks.
We stress that for most practical applications, lexinvariance is far too strong, so these experiments are intended to be illustrative rather than be a recipe for improving state-of-the-art. 
 We discuss potential approaches to integrate the idea of \modelname LM into standard language modeling as a form of regularization, such that the LM assumes some form of partially stable symbol representations. The resulting LM can improve upon a standard language model over some BIG-bench tasks \cite{Srivastava2022BeyondTI}.

\section{\Modelname Language Model}

We define a language model as a probability distribution $p(x_1, \dots, x_n)$ over input token sequences $x_1, \dots, x_n \in \mathcal{V}^n$, where $\mathcal{V}$ is some vocabulary over symbols. 
A language model is \modelname if for all permutations $\pi : \mathcal{V} \to \mathcal{V}$ and for all token sequences $x_1, \dots, x_n \in \mathcal{V}^n$,  $p(x_1, \dots, x_n) = p(\pi(x_1), \dots, \pi(x_n))$. For example, if $\mathcal{V} = \{ \text{a}, \text{b} \}$, then the model should assign the same probability to $aab$ and $bba$.
One example $p$ that satisfies this could simply be 
\vspace{-0.2em}
\begin{gather}
   p(x) =
\begin{cases} 
1/2  & x \in \{aab, bba\}\\
0 & \text{otherwise}
\end{cases}
\end{gather}\label{example}
\vspace{-1.2em}

Can such a \modelname language model predict language well, even though it can only make next token predictions based on the structure of co-occurence and repetition of input tokens in a single context? 

\subsection{Convergence on Language Modeling Performance} 
\label{convergence-theory}

\begin{figure}[tbh]
    \centering
    \vspace{-0.5em}
    \includegraphics[ trim={0 0 0 1cm},clip, width = 0.5\linewidth]{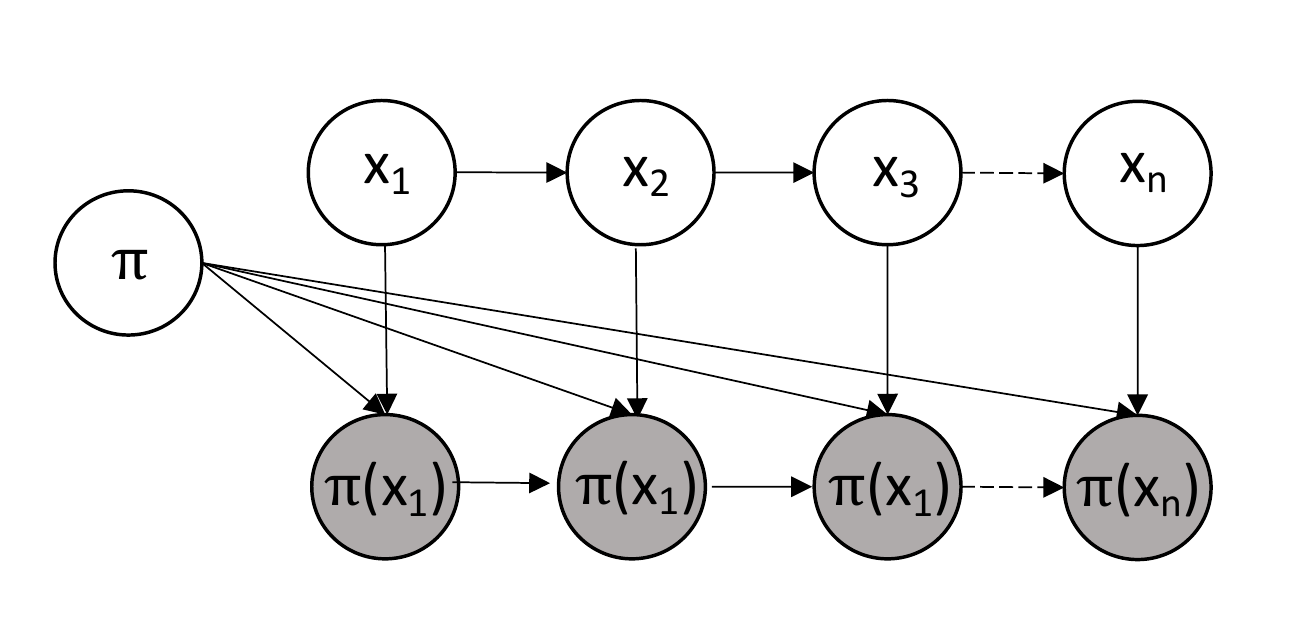} \vspace{-10px}
    \vspace{-0.5em}
    \caption{Probabilistic graphical model for  the \modelname LM associated with the true language distribution $p(x_1, \dots , x_n)$.
    }
    \label{fig:graphical}
    \vspace{-0.5em}
\end{figure}

We show that we can construct a \modelname LM (as shown in Figure \ref{fig:graphical}) to model the true language distribution  faithfully, given a long enough context. The \modelname language model  can essentially infer back the latent permutation $\pi$ as it observes more symbols.

As an intuitive example, suppose  that  $\mathcal{V} = \{ a, b \}$ and the true language only contains two sequences $babbbb$ and $ababab$ (and their prefixes) with even probability.
When given only the first three letters, a \modelname model  can't tell the latent  permutation and can only assign the same probability to $a$ and $b$ for the next letter:
Due to the \modelname property, it assigns the same probability to $p(a|aba)=p(b|bab)$ as well as to $p(b|aba)=p(a|bab)$. Further, $p(a|aba)=p(b|aba)$ because the permutations to prefixes $aba$ and $bab$ are equally probable. In contrast, when considering the prefix $abab$, the fourth letter resolves the ambiguity in possible permutations $\pi$. (Since $baba$ is not in the true language distribution, $\pi$ cannot map $a$ to $b$.) Therefore, the model can correctly predict that $p(a|abab)=1$ and $p(b|abab)=0$.

Formally, for a given language model $p$, we define the associated lexinvariant language model $p'(x_1, \dots, x_n)$ as $\mathop{\mathbb{E}_{\pi}} [p(\pi^{-1}(x_1), \dots, \pi^{-1}(x_n))]$.
Analyzing it, we have the following theorem:

\def\polylog{\operatorname{polylog}}

\begin{theorem}
Let $x_1, \dots, x_n$  be any token sequence generated by an arbitrary language distribution $p$ with an alphabet of size $d$. Let $p'( x_1, \dots, x_{n}) = \mathbb{E}_{\pi} [p( \pi^{-1}(x_1), \dots, \pi^{-1}(x_{n}))]$. 
Then, 
for any $0 < \epsilon, \delta < 1/2$,
    $$\frac1T \sum_{t=1}^T \| p(x_{t} | x_1, \dots, x_{t-1}) - p'( x_{t}| x_1, \dots, x_{t-1}) \|_1  \leq \epsilon  $$
with probability greater than $1-\delta$, 
when $T\geq \frac{d}{\epsilon^4} \; \polylog(d,\frac{1}{\epsilon},\frac{1}{\delta}),$ where the polylogarithmic term hides an absolute constant that is independent of $p$.
\vspace{-0.5em}
\end{theorem}

This theorem says that this associated \modelname language model converges to modeling the true language distribution fairly efficiently---with polynomial rate and near-linear dependence on vocabulary size $d$.  Strikingly, this holds irrespective of the properties of the language distribution $p$ 
\footnote{The convergence rate could be better depending on the language distribution, such as on math and code, where the symbols should have clear functional meaning in context. We explore this empirically in the experiment section.}.
In other words, \textit{a language model can indeed infer the operational meaning of the tokens in context based solely on the structure of the symbols}!

  We give a complete proof of this theorem in Appendix \ref{proof}.  At a high level, this convergence happens because at most timesteps $t$,  the new observation $x_t$ either provides new information about the permutation $\pi$, or $x_t$ has similar likelihood under the permutations that are likely given $x_1,\ldots,x_{t-1}$.  In the simplest case, if the posterior $p(\pi \mid x_1, \dots, x_n)$ concentrates on the correct $\pi$, then we converge to the standard LM.  But even if it doesn't, that means the uncertainty about $\pi$ should not matter for predicting the next token.
  We make this precise by interpreting $p'( x_{t}| x_1, \dots, x_{t-1})$ as performing a multiplicative weights algorithm with the  Hedge strategy of Freund and Schapire \cite{Freund1997ADG}, and then relate the regret bounds to the average KL divergence between the predictions of $p$ and $p'$, and ultimately the average $\mathcal{L}_1$ distance between these predictions.

\subsection{In-context Bayesian Deciphering}

We can see the associated \modelname language model as implicitly learning to approximate an in-context Bayesian deciphering process, i.e. inferring a probability distribution over possible lexical permutations based on seen tokens, with the language modeling prior: 
\begin{align}
\begin{aligned}
    &p'(x_{n+1} |  x_1, \dots, x_{n}) \\
    =& \sum_\pi \underbrace{p( \pi^{-1}(x_{n+1}) | \pi^{-1}(x_1), \dots,  \pi^{-1}(x_{n}) )}_{\text{language modeling}} \underbrace{\mathcal{P}( \pi | x_1, \dots, x_{n})}_{\text{inferring lexical permutation}} 
\end{aligned}
\end{align}

As shown above, $p'$ can be reduced to two parts, where the first part is normal language modeling and the second part is the probability distribution of lexical permutations based on seen tokens. So the \modelname language model is implicitly learning to model $\mathcal{P}( \pi | x_1, \dots, x_{n})$.

We can make this approximate in-context Bayesian deciphering explicit by training a small probe to predict  $\mathcal{P}( \pi | x_1, \dots, x_{n})$ given the internal representation of the \modelname language model. We will show that this indeed recovers $\pi$ reasonably accurately in the experiment section.

\subsection{Constructing a \Modelname Language Model}\label{sec:construction}

We now consider how to construct a \modelname LM in practice. A typical neural language model, such as a Transformer, converts input tokens to continuous vectors using token embedding and then passes these vectors as input to the rest of the neural network.
Thus, the language model $p$ it parameterizes depends on the token embedding $E : \mathcal{V} \to \mathbb R^d$:\looseness=-1
\begin{equation}p(x_1, \dots, x_n) = T(E(x_1), \dots, E(x_n))
\end{equation}
To make a neural LM lexinvariant, we can replace the standard stable token embedding $E$ with a randomized $E$ and take the expectation over $E$. Each token $x \in \mathcal{V}$ has an independent embedding
$E(x) \sim \mathcal{N}(0, \mathcal{I}_d)$, and the language model becomes
\begin{equation}
p(x_1, \dots, x_n) = \mathop{\mathbb{E}} [T(E(x_1), \dots, E(x_n))]
\end{equation}
Since $E \stackrel{d}{=} E \circ \pi$ 
, the right-hand side is the same when $x_i$ are applied with any permutation $\pi$, i.e., for any $x_1, \dots, x_n$:
\begin{align}
\begin{aligned}
    \mathop{\mathbb{E}} [T(E(x_1), \dots, E(x_n))] =  \mathop{\mathbb{E}} [ T(E(\pi(x_1)), \dots, E(\pi(x_n)))],
\end{aligned}
\end{align}
showing that the Transformer with random $E$ is lexinvariant as in Eq. \ref{definition}.
Now we can train this \modelname LM similarly to a standard LM. Concretely, we sample a new $E$ for each training sequence and minimize the standard language modeling loss as in a standard neural LM. Here we are stochastically optimizing a variational lower bound of the standard language modeling loss with this randomized model by taking the expectation to the outside of the loss over log likelihood.
Effectively, the same token gets the same random embedding within each training sequence, but different embedding across training sequences. 

In practice, we focus on training decoder-only Transformers with a next token prediction objective in this work, where the model directly models $p(x_{n+1} | x_1, \dots, x_{n})$ instead of the joint distribution. 
Our definitions and analysis above still hold in general.
The only modification  is that the final readout matrix also needs to be replaced with the same $E$, so that the Transformer can predict the embedding of the next token based on the embeddding of input tokens.

\vspace{-0.5em}
\section{Experiments}
\vspace{-0.3em}

\vspace{-0.3em}
\subsection{Setup}
\vspace{-0.3em}

{\bf Architecture.} For all experiments, we use decoder-only Transformer architecture with T5 relative position bias \cite{T5}. We use models with 150M parameters, with 12 layers, 8 heads, head dimension 128, and MLP dimension 4096. 

{\bf Training.} We use the Adafactor optimizer \cite{DBLP:conf/icml/ShazeerS18}, with a cosine decay learning rate schedule  \cite{Hoffmann2022TrainingCL} from 0.01 to 0.001 based on preliminary experiments. We train the models from scratch for 250K steps on all the settings, with 512 sequence length and 64 batch size. 
We ran all of our experiments on 8 TPU cores. Our models are implemented in JAX \cite{jax2018github}.

{\bf Datasets.} 
For datasets, we mainly use the Pile \cite{Gao2020ThePA}, a large open-source corpus that contains text collected from 22 diverse high-quality sources. We also run experiments on two additional datasets to explore their effects on the behavior of \modelname models:
Wiki-40B \cite{49029}, which contains high quality processed Wikipedia text in 40+ languages, and GitHub (subset of the Pile), which contains code from GitHub repositories with more than 100 stars and less than 1GB files.

\vspace{-0.3em}
\subsection{Convergence to Standard Language Models} \label{sec:convergence}
\vspace{-0.3em}

\begin{figure}[b]
    \centering
    \vspace{-1.7em}
    \begin{minipage}[h]{0.46\linewidth}
    \includegraphics[width=\linewidth]{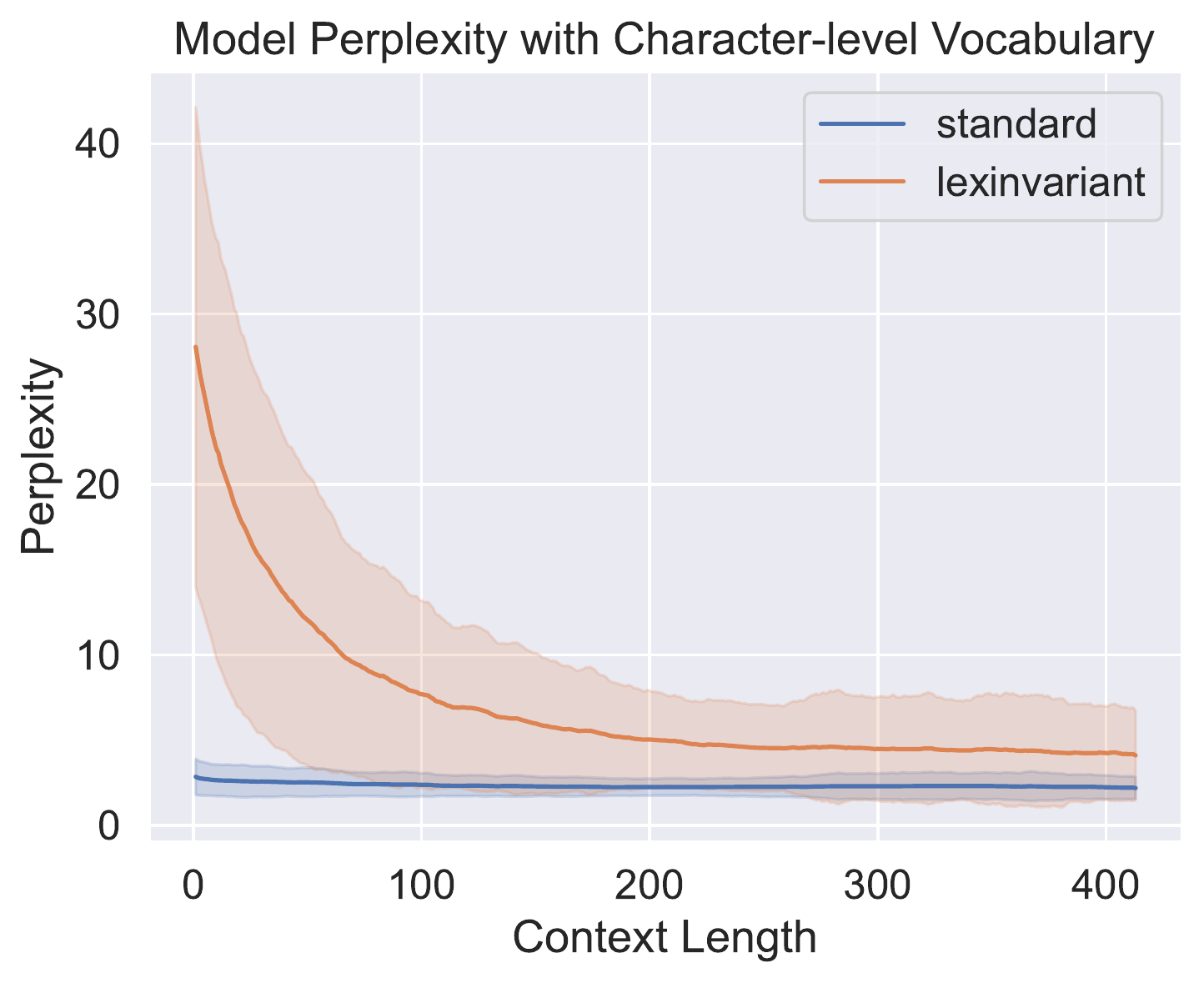}
    \label{fig:convegence1}
    \end{minipage}
    \hfill
    \begin{minipage}[h]{0.50\linewidth}
    \includegraphics[width=\linewidth]{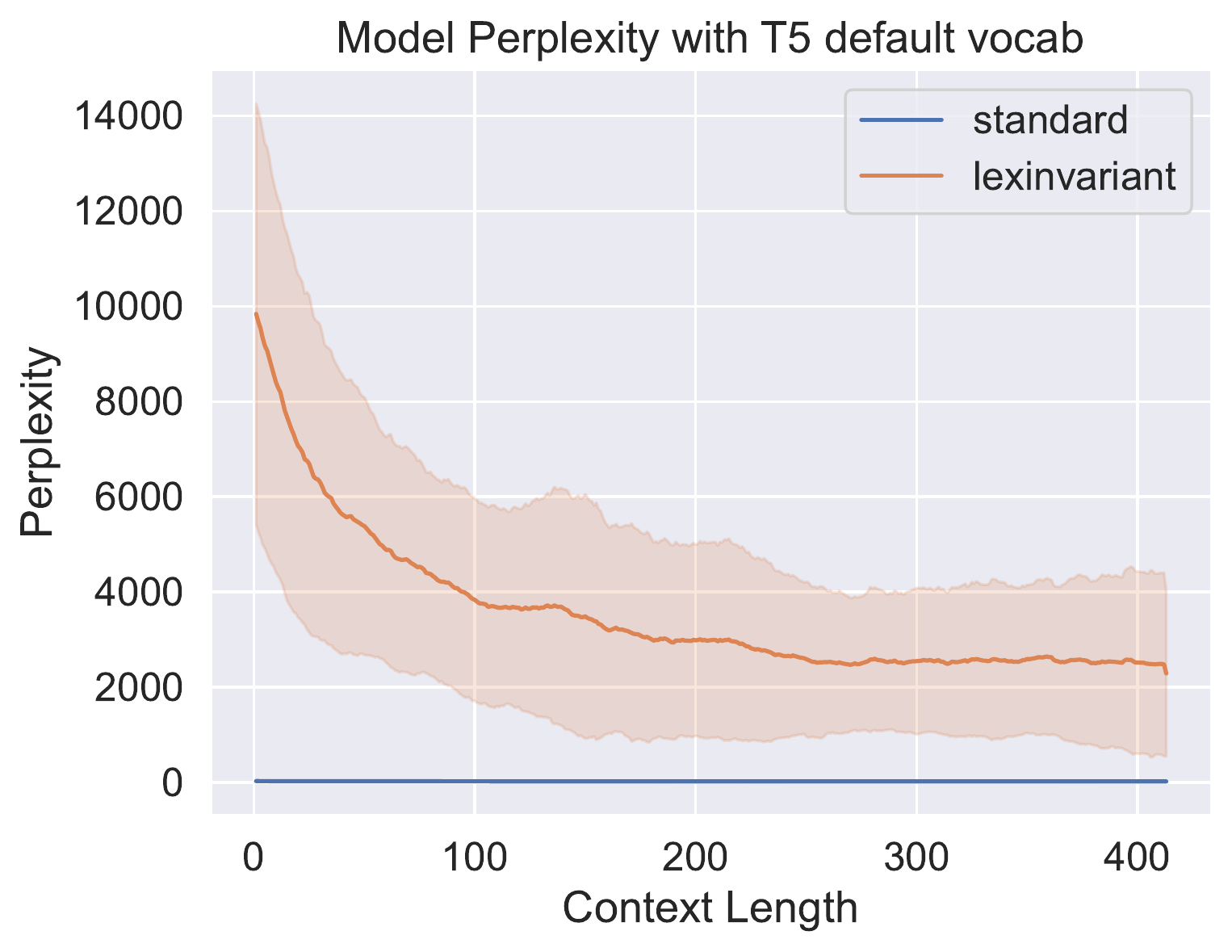}
    \label{fig:convegence2}
    \end{minipage}
    \vspace{-2em}
    \caption{Perplexity over the Pile with character-level vocabulary (left) and T5 default vocab (right).\label{fig:convegence}}
    \vspace{-2em}
\end{figure}

We first show empirically that \modelname LMs can mostly recover the next token prediction performance of standard LMs after a long enough context. As already discussed in section \ref{convergence-theory}, the \modelname LM will theoretically converge to a standard LM as the context becomes long enough to resolve ambiguity. Here we verify this experimentally and show the variation of this convergence across corpora and tokenizations. \looseness=-1

To show this, we train \modelname and standard LMs with both character-level vocabulary (128 ascii characters) and T5 default vocab (32k tokens) over the three datasets. For each model, we measure the perplexity of each token in each sequence w.r.t. context length, smoothed by moving average 
within each sequence, i.e. $ P(x_{i}, \dots, x_{i+k} | x_1, \dots,x_{i} )^{\frac{1}{k}}$ for context length $i$. We set the moving average window $k = 100$. We plot results over 100 sequences. As shown in figure \ref{fig:convegence}, the perplexity gap between \modelname LM and standard LM gradually shrinks as the prefix becomes longer and longer, albeit much more slowly with a larger vocabulary. This makes intuitive sense since a larger vocabulary has more possibilities of permutations and requires many more prefix tokens to disambiguate. For the 32K vocabulary, the 512 context length will only allow the model to see a very small number of tokens, let alone to see tokens more than once. Nonetheless, the model still manages to show the trend of convergence, since even a small number of repetitions can form common patterns in grammar (such as the usage of spaces, punctuation, articles, etc). For the character-level vocabulary, the perplexity gap shrinks from 9X to less than 1X the average perplexity of the standard LM. With a context length of 511, the \modelname LM converges to perplexity 3.38, almost comparable to the perplexity of the standard LM of 2.00.
Additionally, we observe that the gap shrinks significantly faster for models trained over Github than standard English text like Wiki-40B since code is more structured and it is easier to decipher the token permutation. We show the comparison across different datasets in Figure \ref{fig:convegence3} in Appendix.

\vspace{-0.2em}
\subsection{Recovering Substitution Ciphers}
\vspace{-0.2em}

\begin{figure}[b]
\vspace{-2em}
    \centering
    \begin{subfigure}{.3\linewidth}
      \includegraphics[trim={2cm .75cm 2cm 1.5cm},clip,width=\linewidth]{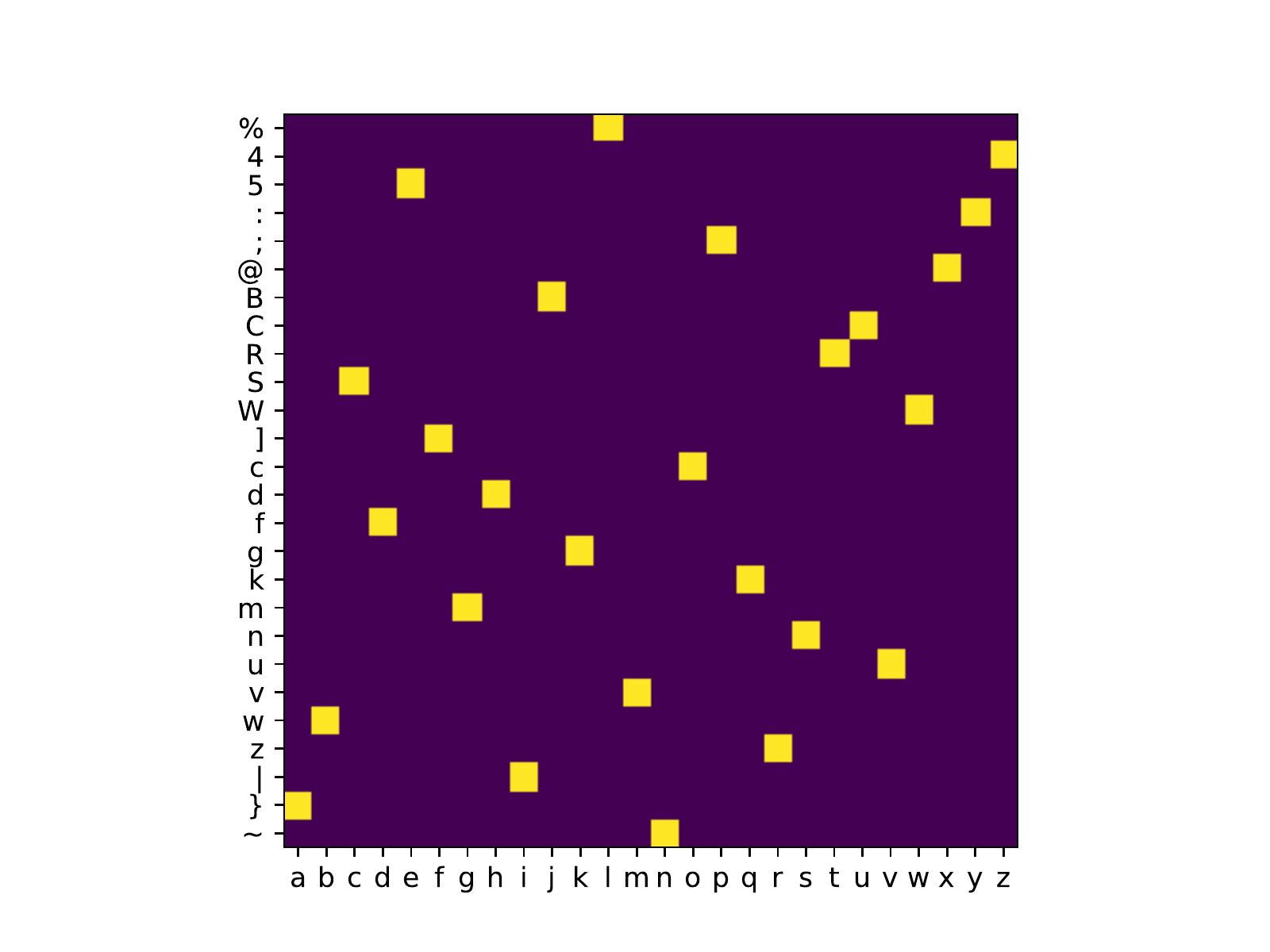}
      \caption{Ground truth}
      \label{fig:gt_and_pred_a}
    \end{subfigure}%
    \begin{subfigure}{0.3\linewidth}
      \includegraphics[trim={2cm .75cm 2cm 1.5cm},clip,width=\linewidth]{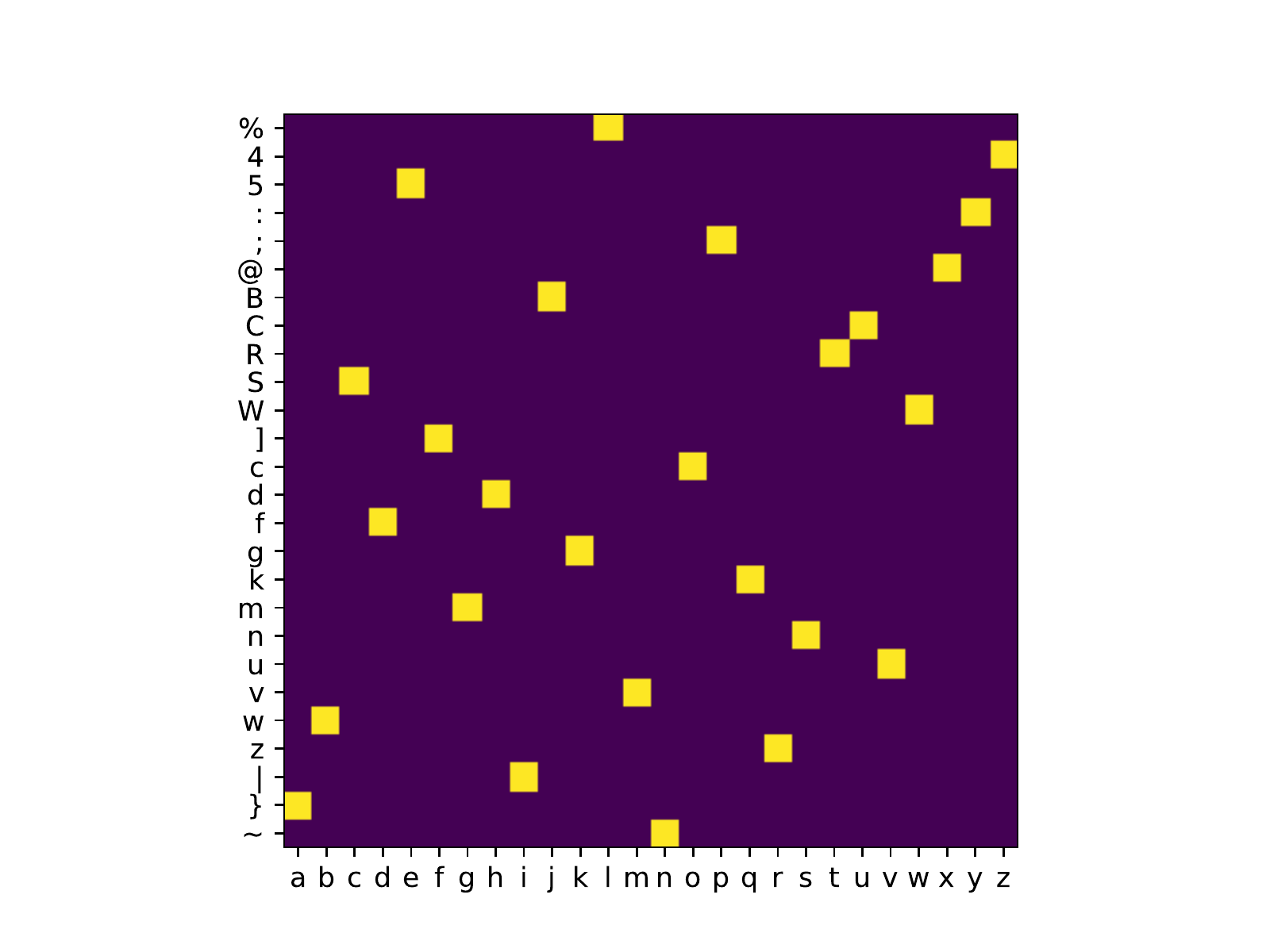}
      \caption{Majority vote prediction}
      \label{fig:gt_and_pred_b}
    \end{subfigure}
    \label{fig:gt_and_pred}
    \begin{subfigure}{0.38\linewidth}
    \centering
    \includegraphics[width=\linewidth]{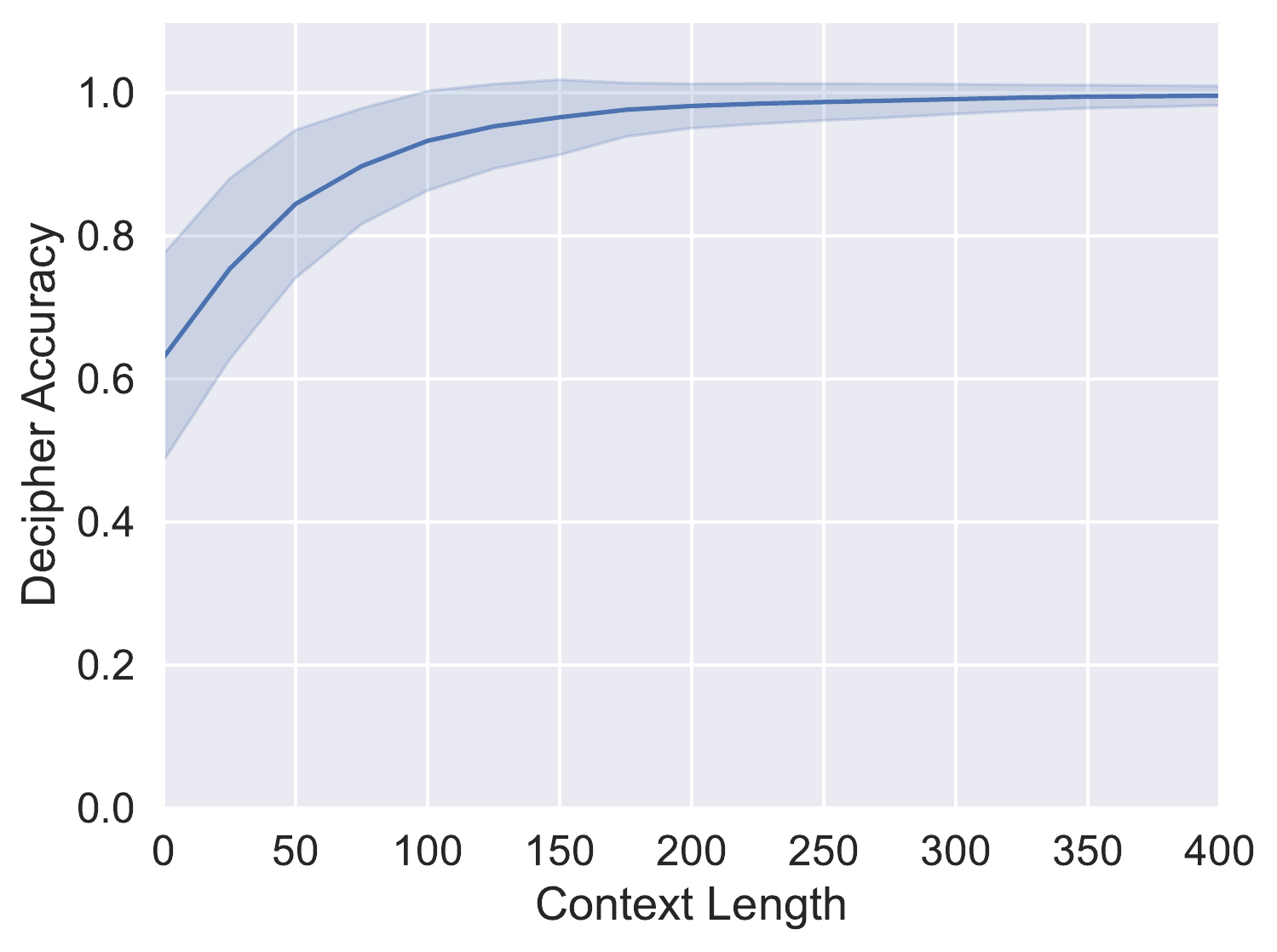}
    \vspace{-2em}
    \caption{
    Cipher key prediction accuracy}
    \label{fig:acc}
    \end{subfigure}
    \vspace{-0.5em}
    \caption{(a) (b): Cipher key matrix, where the vertical axis shows the cipher characters and the horizontal axis shows the deciphered letters. The highlighted entries show the correspondences between cipher characters and the actual letters, e.g. \texttt{\%} deciphers to \texttt{l}.
    (c): Cipher key prediction accuracy, averaged across 1000 input sequences. Context length denotes the start index of the window.}
\end{figure}

Here we show that \modelname LM is implicitly performing Bayesian in-context deciphering by testing its ability to recover cipher keys (e.g. Figure \ref{fig:gt_and_pred_a}) from character-level substitution ciphers, e.g. \texttt{uC; kvR5W 4mfzd @f| Svcgn fw;m uCRmu;;d ]\%\texttildelow\} :fBn}. For the \modelname LM, this cipher text is perceived as the same as \texttt{the quick brown fox jumps over thirteen lazy dogs}, due to the \modelname property. It will then proceed to complete the cipher text with \texttt{ \%d: uC; @f|} with the same probability as it will complete the normal text with \texttt{ and the fox}.

Because of this, we cannot directly read out the distribution of possible cipher keys $\mathcal{P}( \pi | x_1, \dots,  x_{n-1})$ implicitly inferred by the \modelname LM.
To do this, we train a small two-layer MLP probe on top of a frozen trained \modelname LM. 
For each training sequence, we first embed the input sequence with a randomly sampled token embedding $E$ as described in section \ref{sec:construction} and obtain the hidden activation of the final layer generated by the frozen \modelname LM. Then, we pass this activation through the two-layer MLP probe.
Finally, instead of decoding the output activations to classification logits with the same $E$ as in the \modelname LM, we instead use another learnable non-randomized token embedding matrix $E'$ so that the probe can recover the deciphered token with stable token embeddings.
Overall, 
we train the probe jointly with this embedding matrix $E'$ to predict the current token. Effectively, we are training the probe to decipher the current token using the representation provided by the \modelname LM.
We train the probe over the same corpus as the original \modelname LM for 10k steps.
With this probe, we can directly visualize $\mathcal{P}( \pi^{-1}(x_{n}) | x_1, \dots,  x_{n})$ inferred by the \modelname LM, which is effectively one row in the permutation matrix representing $\pi$.

Now we can use this probe to explicitly recover the cipher key. An example ground truth cipher key that we want to recover is shown in Figure \ref{fig:gt_and_pred_a}. Note that although the substitution cipher is only among lowercase letters, the character-level \modelname model we use 
assumes that all permutations among the 128 characters are possible
, making the deciphering even more challenging.

\begin{figure*}[t]
\vspace{-1em}
    \begin{subfigure}{.2\linewidth}
        \includegraphics[trim={2.75cm .75cm 2.75cm 0},clip,width=\linewidth]{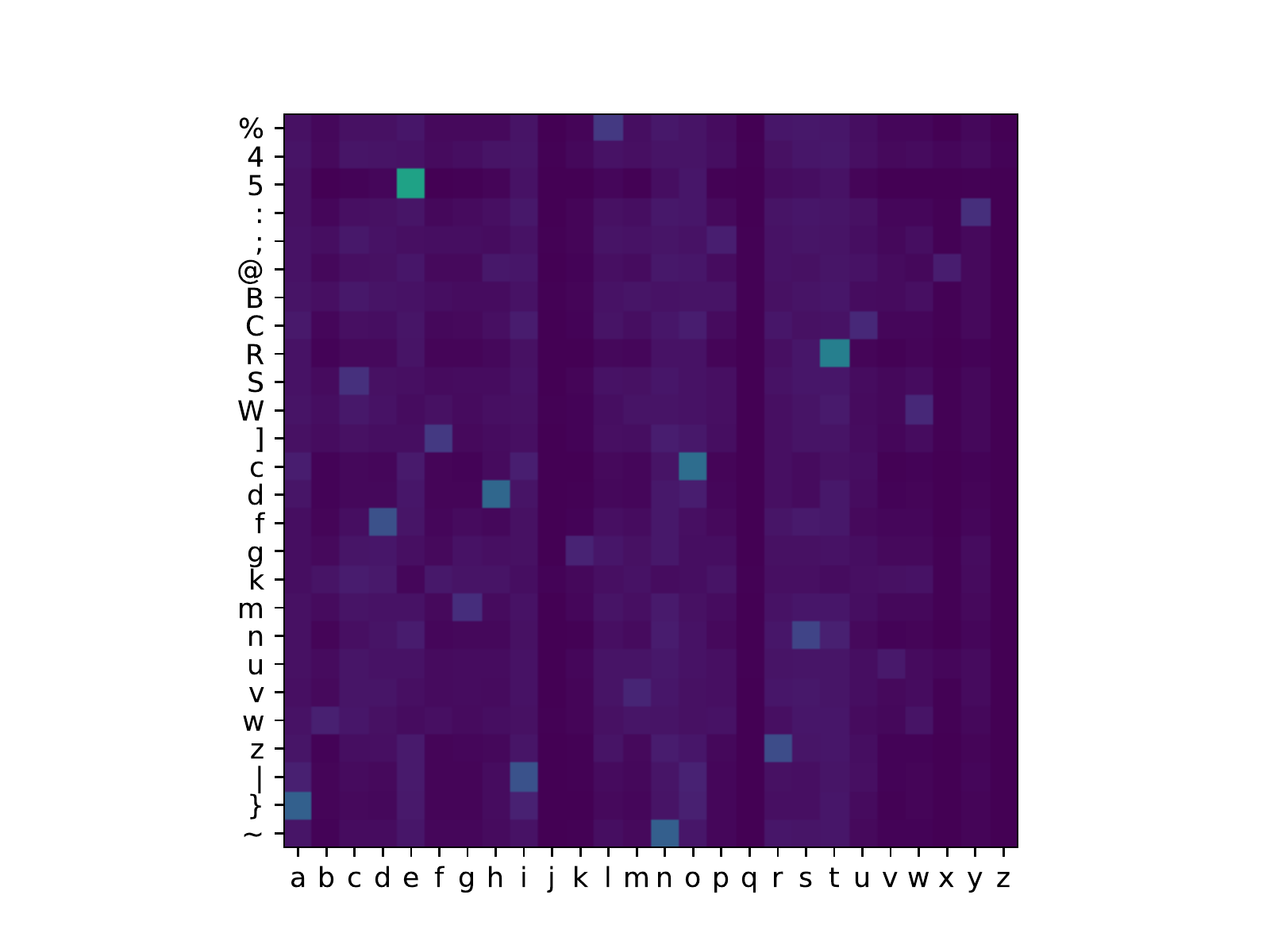}
    \end{subfigure}%
    \begin{subfigure}{.2\linewidth}
        \includegraphics[trim={2.75cm .75cm 2.75cm 0},clip,width=\linewidth]{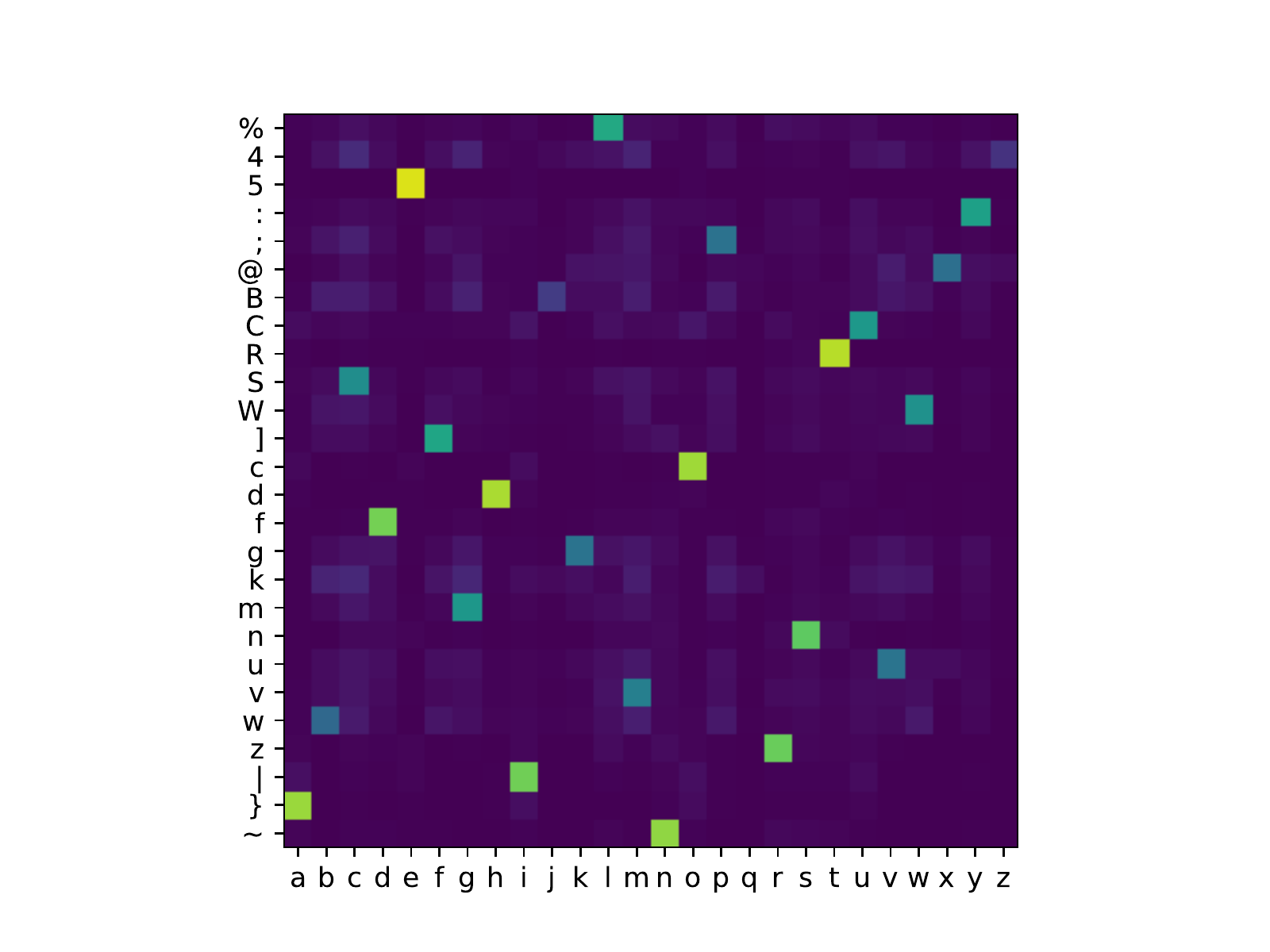}
    \end{subfigure}%
    \begin{subfigure}{.2\linewidth}
        \includegraphics[trim={2.75cm .75cm 2.75cm 0},clip,width=\linewidth]{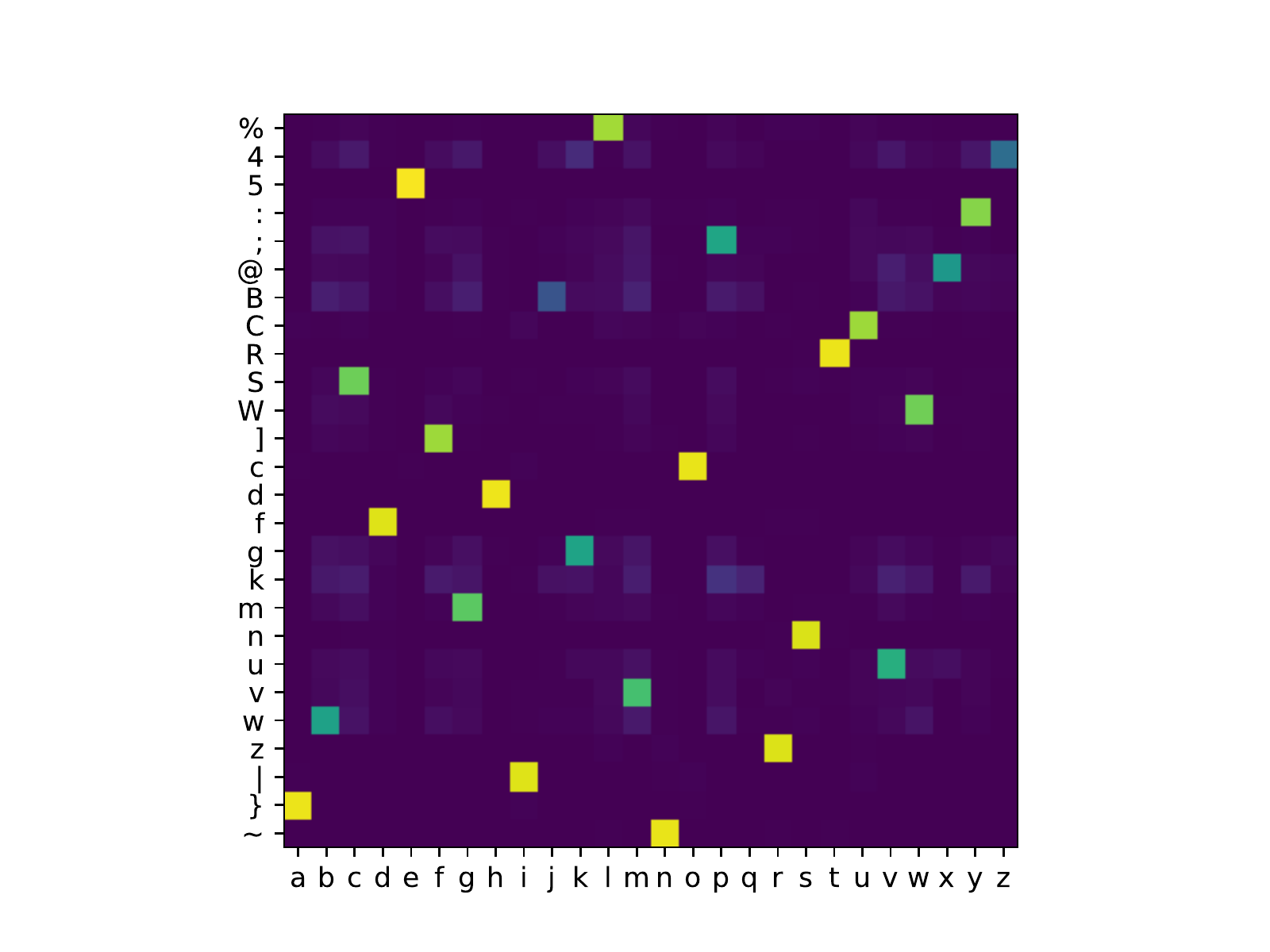}
    \end{subfigure}%
    \begin{subfigure}{.2\linewidth}
        \includegraphics[trim={2.75cm .75cm 2.75cm 0},clip,width=\linewidth]{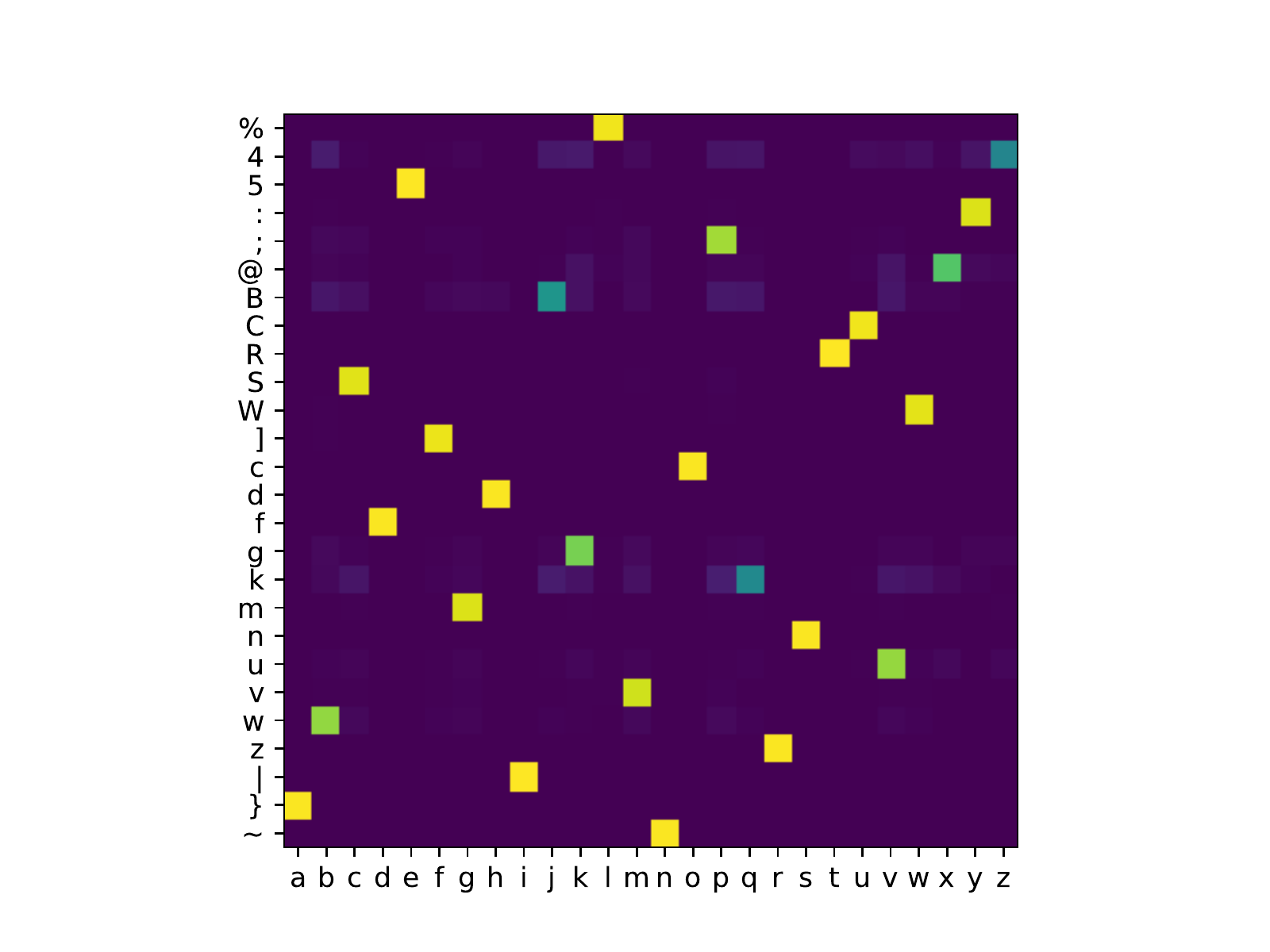}
    \end{subfigure}%
    \begin{subfigure}{.2\linewidth}
        \includegraphics[trim={2.75cm .75cm 2.75cm 0},clip,width=\linewidth]{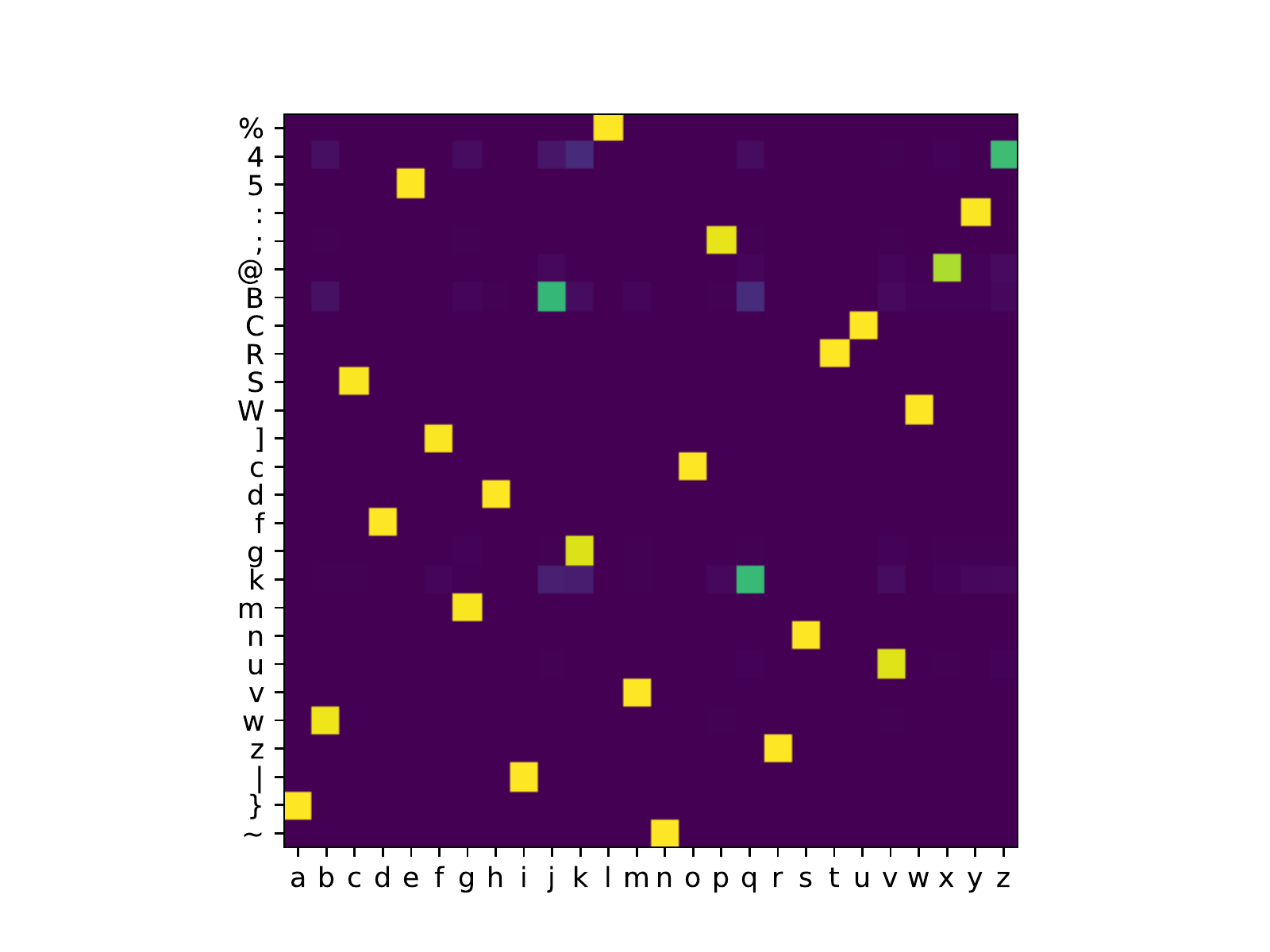}
    \end{subfigure}%
    \vspace{-0.5em}
    \caption{Predicted cipher key for windows of size 50, at indices 0, 50, 100, 200, and 400. Generated using temperature of $T=1$.}
    \label{fig:soft_pred}
    \vspace{-2em}
\end{figure*}

Concretely, we first input ciphertext  through the frozen \modelname LM with the probe to produce a deciphered sequence. We then select a window of size 100 in the middle of the sequence and perform a majority vote over 
the corresponding deciphered tokens of each cipher token seen in this window. This essentially produces a predicted cipher key matrix for each window, and we can measure its precision against the ground truth.
As shown in Figure \ref{fig:acc}, such a cipher key prediction generally has increasingly higher precision as the window is selected later in the context, and it becomes near-perfect by the end of the sequence. Specifically, the cipher key matrix produced by the last window has an average precision of 99.6\% over 1000 input sequences.

Finally, we aggregate over the last window of the 1000 sequences to recover a full cipher key, in case certain letters never appear in the last window  of certain sequences. We again  recover a full cipher key via majority vote. In Figure \ref{fig:gt_and_pred_b}, we show the highly accurate predicted cipher key recovered from ciphertext produced using the example ground truth cipher key in Figure \ref{fig:gt_and_pred_a}.

To perform a more detailed analysis showing the Bayesian deciphering process of the \modelname model, we use the logits of the probe to recover the predicted distribution of the cipher key $\mathcal{P}( \pi | x_1, \dots,  x_{n-1})$. Instead of taking the majority vote of the predicted decipher tokens in the window, we take the mean of logits predicted for each ciphered token. This essentially gives a locally averaged predicted distribution of cipher key matrices. Specifically, the cipher key matrices are generated across windows of 50 characters, and the probabilities are averaged over 1000 input sequences encoded using the same ground truth cipher.
As shown in Figure \ref{fig:soft_pred}, the predicted distribution of cipher key matrix becomes sharper as the prefix becomes longer. 

\vspace{-0.1em}
\subsection{In-context Bayesian Deciphering Examples}\label{sec:decipher}
\vspace{-0.1em}

Here, we show several qualitative examples of in-context Bayesian deciphering. We first show how the \modelname LM maintains uncertainty over possible lexical permutations while iteratively updating them at each index, using examples from a character-level \modelname model. Then, we also show an example of semantic in-context deciphering with a 32K vocabulary \modelname model, where the meaning of a novel word is inferred relative to common words in-context.

\vspace{-0.1em}
\subsubsection{Uncertainty over Lexical Permutations} 
\begin{figure}[b]
\vspace{-1.5em}
    \centering
    \begin{minipage}[h]{0.49\linewidth}
    \begin{subfigure}{\linewidth}
    \includegraphics[trim={0.5cm  3.8cm 0.3cm 4.44cm},clip,width=\linewidth]{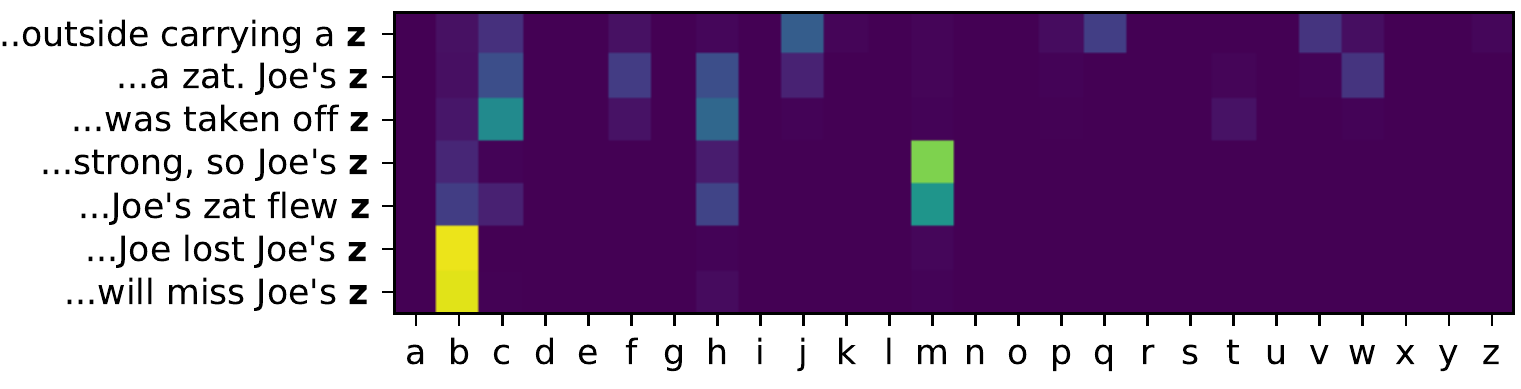}
    \vspace{-5mm}
    \caption{True deciphering: ``\texttt{z}'' $\rightarrow$ ``\texttt{b}'', $T=1$.}
    \label{fig:anec_bat}
    \end{subfigure}

    \begin{subfigure}{\linewidth}
    \includegraphics[trim={0.5cm  3.8cm 0.3cm 4.75cm},clip,width=\linewidth]{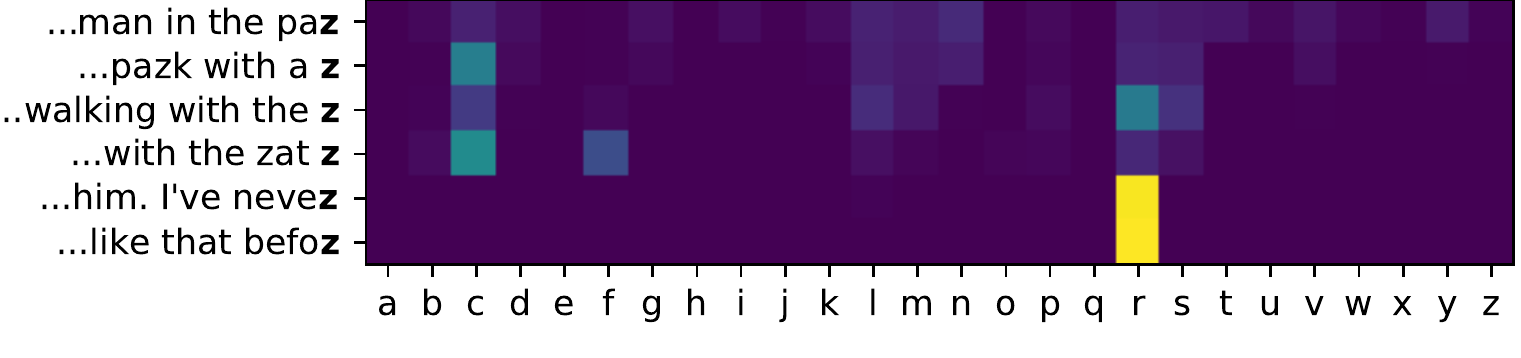}
    \vspace{-6mm}
    \caption{True deciphering: ``\texttt{z}'' $\rightarrow$ ``\texttt{r}'', $T=1$.}
    \label{fig:anec_rat}
    \end{subfigure}
    \end{minipage}
    \begin{minipage}[h]{0.49\linewidth}
    \begin{center}
    \begin{subfigure}{\linewidth}
    \includegraphics[trim={0.5cm 3.8cm 0.3cm 4.75cm},clip,width=\linewidth]{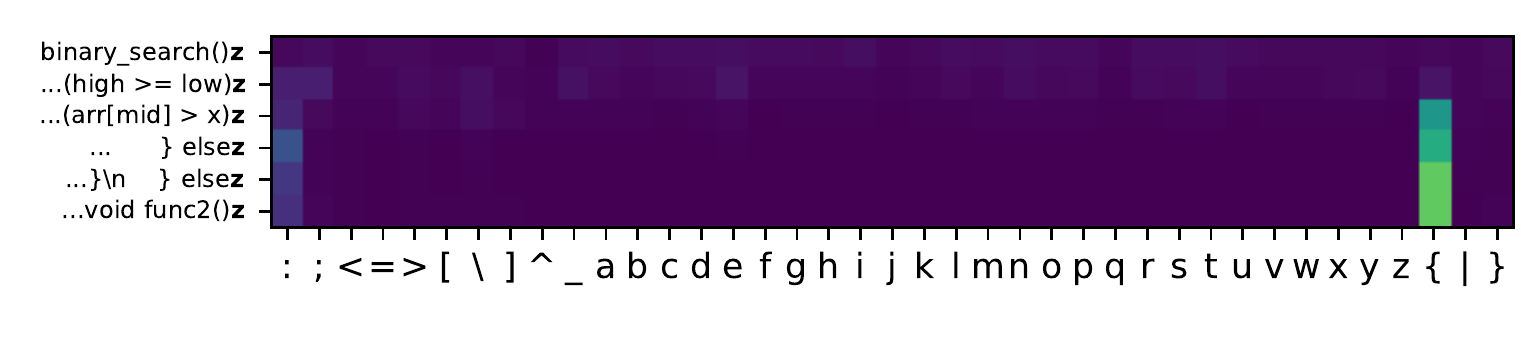}
    \vspace{-7mm}
    \caption{True deciphering: ``\texttt{z}'' $\rightarrow$ ``\{'', $T=2$.}
    \label{fig:anec_java}
    \end{subfigure}

    \begin{subfigure}{\linewidth}
    \includegraphics[trim={0.5cm 3.8cm 0.3cm 4.75cm},clip,width=\linewidth]{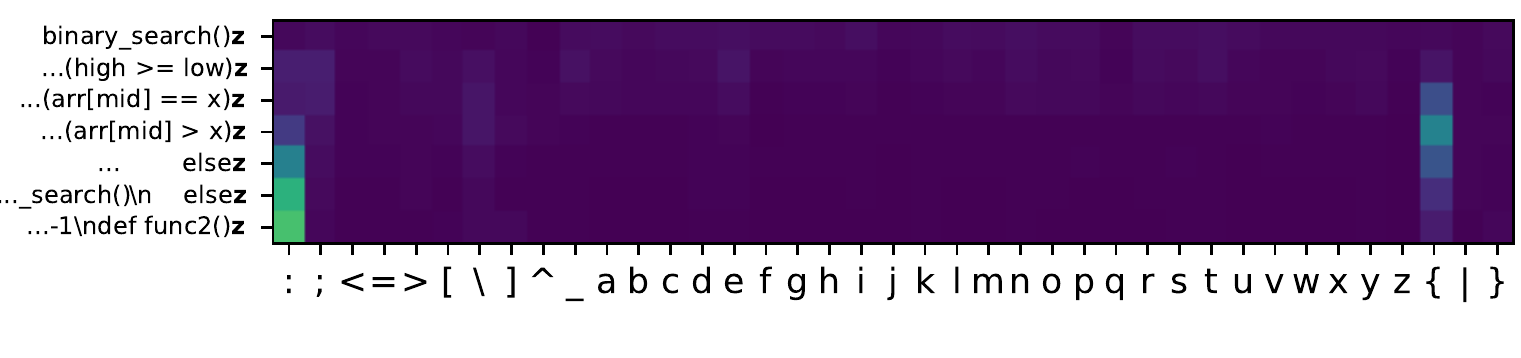}
    \vspace{-7mm}
    \caption{True deciphering: ``\texttt{z}'' $\rightarrow$ ``:'', $T=3$.}
    \label{fig:anec_py}
    \end{subfigure}
    \end{center}
    \end{minipage}
    \vspace{-2mm}
    \caption{Probe predictions for deciphering ``\texttt{z}'' at each occurrence of ``\texttt{z}'' in context.}
    \vspace{-1em}
\end{figure}

\vspace{-0.1em}
 In Figure \ref{fig:anec_bat}, we input the following ciphered sequence to the frozen character-level \modelname LM with the probe: ``\texttt{I saw lots of people in town today, walking and talking around me. I greeted my friend Alice and my classmate Alex. I saw a guy, Joe, walking outside carrying a zat. Joe's zat was taken off zy wind. Today's wind was strong, so Joe's zat flew zackward. Joe lost Joe's zat for good. Joe will miss Joe's zat.}'' 
For each instance of \texttt{z} in the sequence, we display the predicted deciphering of that instance as a row of probabilities across non-cipher letters \texttt{a}-\texttt{z}. 

The \modelname model starts off assuming uniform probability for all possible lexical permutations $\pi$. After seeing more and more text, the \modelname model  quickly realizes that \texttt{z} only has a few main plausible decipherings (\texttt{b}, \texttt{h}, \texttt{c}, \texttt{m}). Eventually, the \modelname model is able to narrow the possibilities down to \texttt{z} maps to \texttt{b} near the end of the sequence. 
The predicted probabilities shift with the seen context accordingly, demonstrating an example of how the predicted cipher key is iteratively updated at each index.

Figure \ref{fig:anec_rat} shows another example with a similar set up, but with text: ``\texttt{I saw a man in the pazk with a zat. The man was walking with the zat zight beside him. I've nevez seen anything like that befoze.}'' 
While context initially suggests that \texttt{z} may be deciphered as \texttt{c}, it becomes clear that \texttt{z} must correspond to \texttt{r} after the appearance of ``\texttt{right}''. The disambiguation is reflected in the depicted probabilities.

In Figure \ref{fig:anec_java} and \ref{fig:anec_py}, we show two deciphering examples over code. We consider two code examples in which it is initially ambiguous whether the character \texttt{z} deciphers to \texttt{:} or \texttt{\{}. The ambiguity is eventually resolved by the use of Python-like or Java-like syntax.

\subsubsection{Semantic Deciphering}\label{sec:semantic}
In addition to character-level deciphering, we show  examples of semantic deciphering with the larger vocabulary of 32k. Although the \modelname LM could not possibly figure out the true lexical permutation among 32k tokens using a small 512 context, it is possible to construct a simple context that repetitively uses simple words so that these words are easier to decipher. Then the \modelname LM can decipher the approximate semantics of the rare symbols relative to other easier-to-decipher words.

One example is the following: given the prompt
\texttt{'crash!' 'aaah!' i looked up from my cup of coffee. 'crash!' - that was the cafe window. and 'aaah!'} [... more text...]\texttt{
what one here is a drink \keys{\return}
- restaurants \keys{\return}- music \keys{\return}- coffee \keys{\return} - father \keys{\return}
the one here that drink is}, where the word \texttt{coffee}, \texttt{music}, and \texttt{father} all only appear once before the question and \texttt{restaurants} appeared 4 times, the model is able to correctly answer that \texttt{coffee} is drinkable. 
See the full example in the appendix.

\vspace{-0.2em}
\subsection{Synthetic Reasoning Tasks}
\vspace{-0.2em}
As discussed in the introduction, lexical flexibility is  correlated with in-context reasoning performance, as demonstrated by existing large LMs.
Thus, 
we study whether the \modelname model also learns in-context reasoning capabilities through the challenging \modelname training.

Specifically, we measure the performance of \modelname models over two pure in-context symbol manipulation tasks: LookUp, where the task is to predict the next token based on the given lookup table, e.g.  \texttt{A->2\keys{\return}C->4\keys{\return}G->5\keys{\return}C->} (should predict \texttt{4} here);
 and Permutation, where the task is to permute an arbitrary subsequence of the given sequence the same way as in the given few demonstrations, e.g. 
\texttt{A 2 C->C A\keys{\return}4 1 D->} (should predict \texttt{D 4} here). 
 In each of the tasks, the symbols are randomly sampled from the vocabulary so that we measure the pure reasoning ability independent from any knowledge of specific words. 
We measure the model performance in terms of generated token accuracy over 1000 examples. The results are shown in Table \ref{tab:syn}. As shown in the table, the \modelname models achieve drastically higher accuracy, with an average of 4X improvement.

\begin{table}[tbh!]
\vspace{-1.2em}
    \centering
    \caption{Accuracy over synthetic reasoning tasks.}
    \begin{tabular}{ll cccc}
    \toprule
          Dataset & Vocab & \multicolumn{2}{c}{LookUp Acc} & \multicolumn{2}{c}{Permutation Acc}\\
         \cmidrule(lr){3-4} \cmidrule(lr){5-6}
          & & Standard & LI & Standard & LI \\
          \midrule
          \multirow{2}{*}{Pile} &char & 48.50 & 91.80 & 27.66 & 59.35 \\
         &32k & 21.45 & 92.10 & 22.84 & 55.63 \\
         \midrule
        \multirow{2}{*}{Wiki-40B} &char & 38.25 & 59.70 & 20.77 & 60.51 \\
        &32k & 8.75 & 59.35 &  9.94 & 50.91 \\
        \midrule
         \multirow{2}{*}{Github} &char & 42.40 & 86.65 & 21.03 & 71.59 \\
          &32k & 4.25 & 80.20 &  8.59 & 67.39 \\
          \bottomrule
    \end{tabular}
    \label{tab:syn}
\vspace{-1.2em}
\end{table}

\subsection{Regularizing Language Models with Lexinvariance}

Although \modelname LM has various interesting properties
, it is not suitable for practical tasks since it would require the context to be extremely long so that all required words and knowledge are defined in the context. Here, we explore how to construct more practical semi-lexinvariant LMs that maintain some properties of \modelname LMs via regularization. We emphasize that this exploration is intended to be illustrative rather than directly improving state-of-the-art.

Instead of using random Gaussian embedding matrices in place of a learned embedding matrix entirely, we can use random embeddings for only some of the tokens in each sequence, while others use the learned embedding. This means that the learned LM assumes that certain tokens have stable meanings but not others, which can be seen as a form of regularization towards lexinvariance.
Specifically, we randomly select tokens to randomize based on a Bernoulli distribution, which can essentially be seen as a form of dropout on token embeddings. On the BIG-bench tasks,  we found that a model with dropout rate $p = 0.2$ for randomization was 25\% more likely to improve performance than to harm performance when evaluated with three shots, relative to a comparably-sized LM, with improvements especially over retrieval type of tasks. See full details in the Appendix \ref{bigbench}.

More broadly, this regularization view could potentially bring the benefit of \modelname LMs to practical applications. For example, the regularization could improve 1) the robustness of LMs by making them less sensitive to adversarial attacks or noise in the input data,
2) generalization across different languages or domains by being less tied to specific lexical items and more prone to learn the shared language structure, and 3) reasoning over more realistic tasks as we have started to explore with BIG-Bench. These areas are promising directions for future work to explore.

\section{Related Work}

 \vspace{-0.2em}
\subsection{Symbol Grounding}
\vspace{-0.2em}
Beyond a modeling choice, the main question of our paper (that being whether an LM can learn language without a stable token representation) is also analogous to the symbol grounding problem: Can meaning be acquired when symbols are not even grounded stably, i.e. they can be mapped to completely random meanings in different sequences? 
It has long been argued by the symbol grounding literature that symbolic representations must be grounded bottom-up in nonsymbolic representations \citep{Harnad1990SymbolGP}, with famous arguments like Searle's Chinese room. And this leads to an ongoing debate on whether LMs can learn meaning purely from large amounts of text, without grounding to any real-world objects \citep{Bender2021OnTD}. Conceptually, \modelname LM is one step further away from physical grounding. We show that they can still infer the meaning of symbols based on lexical structures within the context.

\vspace{-0.2em}
\subsection{Byte-level T5}
\vspace{-0.2em}
There is existing work on absorbing tokenization completely into part of language modeling by using extremely small tokens, such as Byte-level T5 \cite{Xue2021ByT5TA}.
In the extreme, such a model would become closer and closer to \modelname LM, since bytes or bits have almost no stable meaning, so their embeddings are likely not used for prediction. In this paper, we study general \modelname LMs with the \modelname property baked in and without requiring specific tokenizers.

\vspace{-0.2em}
\subsection{Group invariances and Data augmentation}
\vspace{-0.2em}Our implementation of \modelname LMs can be seen as performing a form of very aggressive data augmentation, where we randomize the identity of each token in each sequence. From this perspective, it is somewhat similar to the data recombination in \cite{Jia2016DataRF, Akyrek2020LearningTR} and augmentation of named entities in \cite{Raiman2017GloballyNR}, where certain parts of the sentence are swapped with other words while still maintaining the original grammatical structure. In contrast to these augmentations, the training for our \modelname LMs completely swaps out all parts of the input text.

\vspace{-0.2em}
\subsection{Deciphering Substitution Cipher using LMs}
\vspace{-0.2em}
In general, solving substitution ciphers, where the cipher key is a permutation of the original alphabet, is a NP-hard problem when only having access to LMs that can assign probabilities to sequences \cite{Nuhn2013DeciphermentCI}.
There have been several works focusing on solving substitution ciphers using LMs, including approaches from searching over the permutation space guided by LMs' scores \cite{Hauer2014SolvingSC} to training a seq-to-seq model directly to perform deciphering as translation \cite{Aldarrab2020CanSM}. Although our work does not focus specifically on the task of deciphering substitution ciphers, we show that our \modelname model can efficiently perform in-context deciphering as a byproduct of language modeling.

\vspace{-0.2em}
\subsection{Reasoning}
\vspace{-0.2em}
It has been shown that large language models acquire surprising in-context reasoning capabilities \cite{Brown2020LanguageMA, Liang2022HolisticEO, Srivastava2022BeyondTI}.
Many of them are related to lexical flexibility through training for purely next-token prediction, such as modified arithmetic, data reformatting, and redefining single word etc.
However, LLMs also memorize an enormous amount of knowledge along the way, which is often unnecessary. This work can also be seen as
an exploration of  whether a (semi-)\modelname LM can discount knowledge and prioritize learning the diverse structural reasoning patterns in language, therefore achieving the strong reasoning capability of LLMs with a smaller model.

\section{Conclusion}
In this work, we define and study \modelname language models, which do not have stable embeddings and learn to infer the meaning of symbols in-context. We show several surprising properties of this model theoretically and empirically, including convergence to standard language modeling,  in-context deciphering, and better reasoning capabilities. We also explore a less extreme lexinvariance regularized language model and demonstrate its potential for solving more practical tasks efficiently. 



\bibliography{example_paper}
\bibliographystyle{abbrv}

\newpage
\appendix
\onecolumn

\section{Convergence Proof}\label{proof}

\def\polylog{\operatorname{polylog}}

\begin{theorem}
Let $x_1, \dots, x_n$  be any token sequence generated by an arbitrary language distribution $p$ with an alphabet of size $d$. Let $p'( x_1, \dots, x_{n}) = \mathbb{E}_{\pi} [ p( \pi^{-1}(x_1), \dots, \pi^{-1}(x_{n})) ]$. 
Then, 
for any $0 < \epsilon, \delta < 1/2$,
    $$\frac1T \sum_{t=1}^T \| p(x_{t} | x_1, \dots, x_{t-1}) - p'( x_{t}| x_1, \dots, x_{t-1}) \|_1  \leq \epsilon  $$
with probability greater than $1-\delta$ 
when $T\geq \frac{d}{\epsilon^4} \; \polylog(d,\frac{1}{\epsilon},\frac{1}{\delta}).$
\end{theorem}

\proof
For any desired error $0 < \epsilon < 1/2$ and failure rate $0 < \delta < 1/2$, we will first prove the analogous statement for KL divergence instead of $\mathcal{L}_1$ distance, and then relate a bound on KL divergence back to $\mathcal{L}_1$ distance via Pinsker's inequality.  

Throughout the rest of proof, we will work with a parameter $\epsilon' < O(\frac{\epsilon}{(\log(1/\delta))^{1/4}}) <  
\frac{1}{2}$, and will bound our KL divergence by $\epsilon'$. 

 To prove the bound in terms of KL divergence, it will be useful to ensure to work with a ``smoothed'' version of $p$, which we denote by $\Tilde{p}$, for which every token has some nonzero probability, $\sigma/d$, of appearing at each timestep, for a parameter $\sigma = \delta \epsilon' / T$:
$$\Tilde{p}(x_T | x_1, \dots, x_{T-1})=p(x_T | x_1, \dots, x_{T-1})\left(1-\sigma\right)+ \frac{\sigma}{d}.$$ 
 Similarly, let $\Tilde{p}'( x_1, \dots, x_{n}) = \mathbb{E}_{\pi} [\Tilde{p}^{-1}( \pi(x_1), \dots, \pi^{-1}(x_{n}))]$. We use $\mathcal{\Tilde{P}}$  to denote the probabilities under this change. 
 With probability at least $1- \sigma T \geq 1-\epsilon' \delta \geq 1- \frac{\delta}{2}$, the realized sequence $x_1, \dots, x_n$ drawn under $p$ can be regarded as being drawn from $\Tilde{p}$ (as these distributions can be coupled with this probability).

The key idea is then to show that $\Tilde{p}'(y_{t+1}| y_{1:t})$, where $y_t=\pi^*(x_t)$ for some ground truth $\pi^*$ unknown  to $p'$, is equivalent to using the multiplicative weights algorithm to predict $y_{t+1}$ with the Hedge strategy, with the experts being each possible permutation of the tokens and the cost incurred by each expert being the negative log likelihood of the prediction. We denote 
 $\mathcal{\Tilde{P}}_{\pi'}(y_{1:n}) = \mathcal{\Tilde{P}}( y_{1:n} | \pi = \pi')  = \Tilde{p} ( \pi^{-1}(y_1), \dots, \pi^{-1}(y_{n}))$ and show this in Lemma \ref{equivalence}. 

With this equivalence, we can then bound the difference between the prediction of $p$ and $p'$ as the regret of the multiplicative weights algorithm. Concretely, we show in Lemma \ref{lemma:mweights} that the regret of $p'$ to any expert $\pi$ is bounded as
    $$  \frac1T \sum_t^T \log \frac
    {\mathcal{\Tilde{P}}_{\pi}(y_{t+1} | y_{1:t})}
    {\Tilde{p}'(y_{t+1} | y_{1:t})} 
    \leq 2\epsilon'^2 $$
    for $T\geq \left(4
 \log^2 (\frac{d}\sigma)
\log (d!)\right)/\epsilon'^4$.

We can see $\Tilde{p}$ as the particular expert/permutation $\mathcal{\Tilde{P}}_{I}$. And we can further only consider the special case that $\pi^*$ is also the identity permutation, then the same result holds over $x_t$ and with $\mathcal{\Tilde{P}}_{\pi}$ replaced by $\Tilde{p}$, i.e. $$  \frac1T \sum_t^T \log \frac
    {\Tilde{p}(x_{t+1} | x_{1:t})}
    {\Tilde{p}'(x_{t+1} | x_{1:t})} 
    \leq 2\epsilon'^2 $$

Now we want to convert this bound on regret in terms of log likelihood to KL divergence, and eventually to $\mathcal{L}_1$ distance. To convert it to KL divergence regret, we construct a martingale: $$Z_i = \sum_{t = 1}^i 
        \left(
        D_{KL}(
    \Tilde{p}(x_{t+1} | x_{1:t})
    \|
    \Tilde{p}'(x_{t+1} | x_{1:t})
    ) - \log \frac
        {\Tilde{p}(x_{t+1} | x_{1:t})}
        {\Tilde{p}'(x_{t+1} | x_{1:t})} 
        \right).$$ We verify that this is a martingale in Lemma \ref{martingale}, with differences bounded by $2\log \frac1\sigma$, and bound the probability that $Z_T$ exceeds $b=\log\frac{d}\sigma \sqrt{8T  \log \frac{2}{\delta}} $ via Azuma's inequality Lemma  \ref{lemma:azuma}: with probability $1-\delta/2$, we have that $|Z_T| \leq b$.

Therefore,  we have that with probability at least $1-\delta/2$  
\begin{align*}
    Z_T  = \sum_{t=1}^T \left( D_{KL}(
        \Tilde{p}(x_{t+1} | x_{1:t})
        \|
        \Tilde{p}'(x_{t+1} | x_{1:t})
    ) 
    - \log \frac
        {\Tilde{p}(x_{t+1} | x_{1:t})}
        {\Tilde{p}'(x_{t+1} | x_{1:t})} 
        \right) &\leq b\\
\sum_{t=1}^T D_{KL}(
        \Tilde{p}(x_{t+1} | x_{1:t})
        \|
        \Tilde{p}'(x_{t+1} | x_{1:t})
    ) 
        &\leq \sum_{t=1}^T \left(\log \frac
        {\Tilde{p}(x_{t+1} | x_{1:t})}
        {\Tilde{p}'(x_{t+1} | x_{1:t})} \right) + b
\end{align*}

Putting this all together, since $\frac1T \sum_t^T \log \frac
    {\Tilde{p}_{i}(x_{t+1} | x_{1:t})}
    {\Tilde{p}'(x_{t+1} | x_{1:t})} 
    \leq 2\epsilon'^2$ for $T\geq \left(4
 \log^2 (\frac{d}\sigma)
\log (d!)\right)/\epsilon'^4$, we have the following:

\begin{align*}
\sum_1^T D_{KL}(
    \Tilde{p}(x_{t+1} | x_{1:t})    \|        \Tilde{p}'(x_{t+1} | x_{1:t})
) 
&\leq 2\epsilon'^2 T + b.
\end{align*}

We now convert our bound on KL divergence to a bound on $\mathcal{L}_1$ distance via Pinsker's inequality:
$$\| \Tilde{p}(x_{t+1} |x_{1:t} ) - \Tilde{p}'(x_{t+1} | x_{1:t}) \|_1 \leq \sqrt{\frac12 D_{KL}(\Tilde{p}(x_{t+1} | x_{1:t})||\Tilde{p}'(x_{t+1} | x_{1:t}))}.$$

Further, at any given $x_t$, the difference between 
the redistributed probability distribution $\Tilde{p}$ and a unmodified probability distribution $p$ is at most $\sigma$, so

\begin{align*}
    \| p(x_{t+1} |x_{1:t} ) - p'(x_{t+1} | x_{1:t}) \|_1 
    &\leq \| \Tilde{p}(x_{t+1} |x_{1:t} ) - \Tilde{p}'(x_{t+1} | x_{1:t}) \|_1  + 2\sigma.
\end{align*}

We are interested in the average $\mathcal{L}_1$ across time steps:
\begin{align*}
    \frac1T \sum_{t=1}^T \| p(x_{t+1} |x_{1:t} ) - p'(x_{t+1} | x_{1:t}) \|_1 
    &\leq \frac1T \sum_{t=1}^T \left( \| \Tilde{p}(x_{t+1} |x_{1:t} ) - \Tilde{p}'(x_{t+1} | x_{1:t}) \|_1 + 2\sigma\right)\\
    &\leq \frac1T \sum_{t=1}^T  \sqrt{\frac12 D_{KL}( \Tilde{p}(x_{t+1} | x_{1:t})  \|  \Tilde{p}'(x_{t+1} |x_{1:t}))}    + 2\sigma\\
    &\leq \frac1T \sqrt{ T  \sum_{t=1}^T \frac{1}{2} { D_{KL}( \Tilde{p}(x_{t+1} | x_{1:t})  \|  \Tilde{p}'(x_{t+1} |x_{1:t}))}}   + 2\sigma,\\
\end{align*}
where in the last inequality we applied Cauchy–Schwarz. Hence for $T\geq \left(4
 \log^2 \frac{d}\sigma
\log (d!)\right)/\epsilon'^4$, 
\begin{align*}
    \frac1T \sum_{t=1}^T \| p(x_{t+1} | x_1, \dots, x_{t}) - p'(x_{t+1} | x_1, \dots, x_{t}) \|_1 
    &\leq  \frac1T \sqrt{ \frac{T}{2} (2\epsilon'^2 T + b)} + 2\sigma\\
        &\leq \sqrt{\epsilon'^2+\frac{b}{2T}} + 2\sigma.
\end{align*}

Simplifying this for $b=\log\frac{d}\sigma \sqrt{8T  \log \frac{2}{\delta}},$  $T\geq \left(4
 \log^2 (\frac{d}{\sigma})
\log (d!)\right)/\epsilon'^4$ and $\sigma = \epsilon' \delta / T$, we have 
\begin{align*}
    \frac1T \sum_{t=1}^T \| p(x_{t+1} | x_1, \dots, x_{t}) - p'(x_{t+1} | x_1, \dots, x_{t}) \|_1 
    &\leq  \sqrt{\epsilon'^2 + \frac{\sqrt{2 \log \frac{2}{\delta}}}{\sqrt{\log (d!)}} \epsilon'^2} + \frac{2\epsilon' \delta }{ T}\\
    &\leq \epsilon'(\frac{2 \delta }{ T} + \sqrt{1 + \sqrt{\frac{2 \log \frac{2}{\delta}}{\log (d!)}}})\\
    &\leq \epsilon'(1 + \sqrt{1 + \sqrt{2 \log \frac{2}{\delta}}})  \leq \epsilon' 2\sqrt{2} (2 \log \frac{2}{\delta})^{1/4}. 
\end{align*}

We can bound this average $L_1$ error by $\epsilon$ if we set $\epsilon' = \frac{\epsilon}{2\sqrt{2} (2 \log \frac{2}{\delta})^{1/4} } < \frac{1}{2}$, in which case our condition that $T\geq \left(4
 \log^2 (\frac{dT}{\delta \epsilon'})
\log (d!)\right)/\epsilon'^4$ becomes  $T\geq \left(512 \log \frac{2}{\delta}
 \log^2 \frac{dT}{\delta \epsilon'}
\log (d!)\right)/\epsilon^4$.   The theorem now follows by simplifying this expression.  Since $\log \frac{2}{\delta} \leq 2\log \frac{1}{\delta}$, and $\log (d!) \leq d\log (d)$, we can relax the condition on $T$ as $$T \geq \left(1024 \log \frac{1}{\delta}
 \log^2 (\frac{d}{\delta \epsilon'})\log^2 (T)
d\log (d)\right)/\epsilon^4 = \log^2 (T) \frac{d}{\epsilon^4} \; polylog(d,\frac{1}{\epsilon},\frac{1}{\delta})$$

To remove the $\log^2 T$ from the right side, note that for any $W>10$, if T > 10 $W \log^2 W$, then $T > W log^2 T$, yielding the further relaxed the condition on $T$ as
$$T \geq \frac{d}{\epsilon^4} \; polylog(d,\frac{1}{\epsilon},\frac{1}{\delta}).$$
\qed

\begin{lemma} \label{equivalence}
Consider an arbitrary ground truth permutation $\pi^*$. For all time steps $t\in[1,n]$, let $y_t=\pi^*(x_t)$. Consider the online prediction game of predicting $y_{t+1}$ at each time step given previous observation $y_{1:t}$ without knowing $\pi^*$ but knowing $\Tilde{p}$. Then, $\Tilde{p}'(y_{t+1}| y_{1:t})$  is equivalent to the multiplicative weights algorithm's prediction of $y_{t+1}$ with the  Hedge strategy of Freund and Schapire \cite{Freund1997ADG}, where it 
\begin{itemize}
    \item Considers $d!$ experts corresponding to guessing each permutation $\pi'$ is the ground truth permutation.
    \item Maintains a weight $w_{\pi'}^{(t)}$ for each expert at time step $t$, and the weights are initially as $\mathcal{\Tilde{P}}(\pi)$.
    \item Picks a distribution across experts $p_{\pi'}^{(t)}=\frac{w_{\pi'}^{(t)}}{\Phi^{(t)}}$ where $\Phi^{(t)}=\sum_{j} w_{j}^{(t)}$.
    \item Produces prediction of $y_{t+1}$ as  $ \sum_{\pi'}p_{\pi'}^{(t)}\mathcal{\Tilde{P}}_{\pi'}(y_{t+1} | y_{1:t}) $
    \item Receives a cost vector of $m^{(t)}_{\pi'}=-\frac{1}{\epsilon} \log{\mathcal{\Tilde{P}}_{\pi'}(y_{t+1} | y_{1:t})}$.
     \item Updates the weights $w_i^{(t+1)}=w_i^{(t)} \exp (-\epsilon m_i^{(t)})$ and repeat
\end{itemize}
\end{lemma}

\proof We can first see that $p_{\pi'}^{(t)} = \mathcal{\Tilde{P}}(\pi' | y_{1:t})$ by induction:

Base case: $p_{\pi'}^{(0)} = \mathcal{\Tilde{P}}(\pi)$ by assumption.

Inductive Case:

With the cost vector as 
$m^{(t-1)}_{\pi'}=-\frac{1}{\epsilon} \log{\mathcal{\Tilde{P}}_{\pi'}(y_{t} | y_{1:t-1})}$,
the update at step $t$ is $w_{\pi'}^{(t)}=w_{\pi'}^{(t-1)}  \mathcal{\Tilde{P}}_{\pi'}(y_{t} | y_{1:t-1})$.
Therefore, the probability over any particular expert $\pi'$ is
\begin{align*}\label{mul}
    p_{\pi'}^{(t)}&=\frac{w_{\pi'}^{(t)}}{\Phi^{(t)}}\\
    &=\frac{w_{\pi'}^{(t-1)}  \mathcal{\Tilde{P}}_{\pi'}(y_{t} | y_{1:t-1})}{\sum_j w_{j}^{(t-1)}  \mathcal{\Tilde{P}}_{j}(y_{t} | y_{1:t-1})}\\
    &=\frac{p_{\pi'}^{(t-1)}\Phi^{(t-1)}  \mathcal{\Tilde{P}}_{\pi'}(y_{t} | y_{1:t-1})}{\sum_j p_{j}^{(t-1)}\Phi^{(t-1)}  \mathcal{\Tilde{P}}_{j}(y_{t} | y_{1:t-1})}\\
    &=\frac{
    p_{\pi'}^{(t-1)}  
    \mathcal{\Tilde{P}}_{\pi'}(y_{t} | y_{1:t-1})}{\sum_j p_{j}^{(t-1)} \mathcal{\Tilde{P}}_{j}(y_{t} | y_{1:t-1})}\\
\end{align*} 

This is equivalent to the update given by Bayes' rule when plugging in $p_{\pi'}^{(t)} = \mathcal{\Tilde{P}}(\pi' | y_{1:t})$ :
\begin{align*}
    \mathcal{\Tilde{P}}(\pi' | y_{1:t}) &= \frac{
    \mathcal{\Tilde{P}}(\pi' | y_{1:t-1})
    \mathcal{\Tilde{P}}_{\pi'}(y_{t} | y_{1:t-1})}
    {\mathcal{\Tilde{P}}(y_{t} | y_{1:t-1})}
\end{align*}

So we can conclude that $p_{\pi'}^{(t)} = \mathcal{\Tilde{P}}(\pi' | y_{1:t})$, i.e. the process of updating the probability distribution across experts within the prediction game is equivalent to the process of the language model updating the probabilities $\mathcal{\Tilde{P}}(\pi' | y_{1:t+1}) $ across permutations $\pi'$. And this means that the algorithm's prediction $\sum_{\pi'}p_{\pi'}^{(t)}\mathcal{\Tilde{P}}_{\pi'}(y_{t+1} | y_{1:t}) = \sum_{\pi'}  \mathcal{\Tilde{P}}(\pi' | y_{1:t}) \mathcal{\Tilde{P}}_{\pi'}(y_{t+1} | y_{1:t}) = \mathcal{\Tilde{P}}(y_{t+1}| y_{1:t}) = 
\Tilde{p}'(y_{t+1}| y_{1:t}) $  
\qed

\begin{lemma}
    \label{lemma:mweights}
    When using the Hedge strategy for the multiplicative weights algorithm, the average difference between the weighted distribution across experts and any particular expert $\pi$ is bounded as
    $$  \frac1T \sum_t^T \log \frac
    {\mathcal{\Tilde{P}}_{\pi}(y_{t+1} | y_{1:t})}
    {\Tilde{p}'(y_{t+1} | y_{1:t})} 
    \leq 2\epsilon^2 $$
    for $\epsilon\leq1$ and for $T\geq \left(4
 \log^2 \left(\frac{d}\sigma \right)
\log (d!)\right)/\epsilon^4$.
\end{lemma}

\proof
Consider an arbitrary expert $\pi$.

We first show that the cost vectors are bounded by $\rho= -\frac{1}{\epsilon} \log \frac\sigma{d}$:
Recall we defined
$m^{(t)}_{\pi}=-\frac{1}{\epsilon} \log{\mathcal{\Tilde{P}}_{\pi}(y_{t+1} | y_{1:t})}$. By the definition of our redistributed probability distribution, at time step $t\in[1,\dots,T]$, 
\begin{align*}
    \frac\sigma{d} &\leq \mathcal{\Tilde{P}}_{\pi}(y_{t+1} | y_{1:t}) \leq 1\\
    \log  \frac\sigma{d}  &\leq \log\mathcal{\Tilde{P}}_{\pi}(y_{t+1} | y_{1:t}) \leq 0 \\
    0 &\leq m^{(t)}_{\pi} \leq -\frac{1}{\epsilon} \log \frac\sigma{d}\\
    0 &\leq m^{(t)}_{\pi} \leq -\frac{1}{\epsilon} \log \frac\sigma{d}.
\end{align*}

By corollary 16.3 in \cite{MW}, 
if we have cost vectors $m^{(t)}\in[-\rho, \rho]^{d!}$, then for time $T\geq (4\rho^2\log (d!))/\epsilon^2$ where $\epsilon\leq1$, 
\begin{align*}
    \frac1T\sum_t^T p^{(t)}\cdot m^{(t)} \leq \frac1T \sum_t^T m_{\pi}^{(t)}+2\epsilon.
\end{align*}

Note that we can simplify $T\geq \left(4
 \log^2 \left(\frac{d}\sigma\right)
\log (d!)\right)/\epsilon^4$.

We can now bound 
\begin{align*}
    \frac1T\sum_t^T \left( p^{(t)}\cdot m^{(t)} -   m_{\pi}^{(t)}\right) &\leq 2\epsilon\\    
    \frac1T\sum_t^T 
    \left(
    \sum_{\pi'}
    p^{(t)}_{\pi'} m^{(t)}_{\pi'} - m_{\pi}^{(t)}\right) 
    &\leq  2\epsilon\\
    \frac1T\sum_t^T 
    \left(
    \sum_{\pi'}
    \mathcal{\Tilde{P}}(\pi' | y_{1:t})  
    \left(-\frac{1}{\epsilon} \log{\mathcal{\Tilde{P}}_{\pi'}(y_{t+1} | y_{1:t})}\right) 
    - \left(-\frac{1}{\epsilon} \log{\mathcal{\Tilde{P}}_{\pi}(y_{t+1} | y_{1:t})}\right) 
    \right)
    &\leq 2\epsilon\\
    \frac{1}{\epsilon T}\sum_t^T 
    \sum_{\pi'}
    \left(
    \mathcal{\Tilde{P}}(\pi' | y_{1:t})  
    \left(\log{\mathcal{\Tilde{P}}_{\pi}(y_{t+1} | y_{1:t})} - \log{\mathcal{\Tilde{P}}_{\pi'}(y_{t+1} | y_{1:t})}\right) 
    \right)
    &\leq 2\epsilon\\
    \frac1T \sum_t^T \mathbb{E}_{\pi'} \log \frac
    {\mathcal{\Tilde{P}}_{\pi}(y_{t+1} | y_{1:t})}
    {\mathcal{\Tilde{P}}_{\pi'}(y_{t+1} | y_{1:t})} 
    &\leq 2\epsilon^2
\end{align*}

By Jensen's inequality, we also have that
\begin{align*}
    \frac1T \sum_t^T \log \frac
    {\mathcal{\Tilde{P}}_{\pi}(y_{t+1} | y_{1:t})}
    {\mathbb{E}_{\pi'} 
 \mathcal{\Tilde{P}}_{\pi'}(y_{t+1} | y_{1:t})} 
    &\leq 2\epsilon^2 \\
    \frac1T \sum_t^T \log \frac
    {\mathcal{\Tilde{P}}_{\pi}(y_{t+1} | y_{1:t})}
    {\Tilde{p}'(y_{t+1} | y_{1:t})} 
    &\leq 2\epsilon^2 
\end{align*}

\qed

\begin{lemma} \label{martingale}
    Let
    \begin{align*}
        Z_i &= \sum_{t=1}^i
        \left(
        D_{KL}(
    \mathcal{\Tilde{P}}_I(x_{t+1} | x_{1:t})
    \|
    \mathcal{\Tilde{P}}(x_{t+1} | x_{1:t})
    ) - \log \frac
        {\mathcal{\Tilde{P}}_I(x_{t+1} | x_{1:t})}
        {\mathcal{\Tilde{P}}(x_{t+1} | x_{1:t})} 
        \right)
    \end{align*}
$Z_i$ is a martingale.
\end{lemma}
\proof

Consider 
\begin{align*}
    \mathbb{E}_{x_{i+1}\sim \mathcal{\Tilde{P}}_I} \left[Z_i\right]
    &=
    \mathbb{E}_{x_{i+1}\sim \mathcal{\Tilde{P}}_I} \left[
    \sum_{t=1}^i
        \left(
        D_{KL}(
    \mathcal{\Tilde{P}}_I(x_{t+1} | x_{1:t})
    \|
    \mathcal{\Tilde{P}}(x_{t+1} | x_{1:t})
    ) - \log \frac
        {\mathcal{\Tilde{P}}_I(x_{t+1} | x_{1:t})}
        {\mathcal{\Tilde{P}}(x_{t+1} | x_{1:t})} 
        \right)
    \right] \\ 
&= \mathbb{E}_{x_{i+1}\sim \mathcal{\Tilde{P}}_I} \left[
        D_{KL}(
    \mathcal{\Tilde{P}}_I(x_{i+1} | x_{1:i})
    \|
    \mathcal{\Tilde{P}}(x_{i+1} | x_{1:i})
    ) - \log \frac
        {\mathcal{\Tilde{P}}_I(x_{i+1} | x_{1:i})}
        {\mathcal{\Tilde{P}}(x_{i+1} | x_{1:i})}  + Z_{i-1}
    \right]
\end{align*}
Observe that $Z_{i-1}$ has no dependence on $x_{i+1}$.

\begin{align*}
\mathbb{E}_{x_{i+1}\sim \mathcal{\Tilde{P}}_I} \left[Z_i\right]
    &= \mathbb{E}_{x_{i+1}\sim \mathcal{\Tilde{P}}_I} \left[
    \mathbb{E}_{x_{i+1}\sim \mathcal{\Tilde{P}}_I} \log \frac
        {\mathcal{\Tilde{P}}_I(x_{i+1} | x_{1:i})}
        {\mathcal{\Tilde{P}}(x_{i+1} | x_{1:i})} \right]
    - \mathbb{E}_{x_{i+1}\sim \mathcal{\Tilde{P}}_I} \left[ \log \frac
        {\mathcal{\Tilde{P}}_I(x_{i+1} | x_{1:i})}
        {\mathcal{\Tilde{P}}(x_{i+1} | x_{1:i})} 
    \right] + Z_{i-1}\\
    &= Z_{i-1}
\end{align*}

Therefore, $Z_i$ is a martingale.\qed

\begin{lemma} 
$|Z_i-Z_{i-1}|\leq c_i$
where $c_i = 2 |\log\frac{d}{\sigma}|$
\end{lemma}
\proof
We have
\begin{align*}
    |Z_i-Z_{i-1}| &= \left|
        D_{KL}(
    \mathcal{\Tilde{P}}_I(x_{i+1} | x_{1:i})
    \|
    \mathcal{\Tilde{P}}(x_{i+1} | x_{1:i})
    ) - \log \frac
        {\mathcal{\Tilde{P}}_I(x_{i+1} | x_{1:i})}
        {\mathcal{\Tilde{P}}(x_{i+1} | x_{1:i})} \right|
\end{align*}

In our redistributed probability distribution $\mathcal{\Tilde{P}}$, we have $\frac\sigma{d} \leq \mathcal{\Tilde{P}}_{\pi}(x_i  | x_{1:i-1}) \leq 1$ for any $\pi$ at any time $i$.
Therefore,
\begin{align*}
 \log\frac\sigma{d} 
 \leq 
\log \frac
        {\mathcal{\Tilde{P}}_I(x_{i+1} | x_{1:i})}
        {\mathcal{\Tilde{P}}(x_{i+1} | x_{1:i})}
 \leq \log \frac{d}{\sigma}.
\end{align*}

Also, we can find an upper bound for the KL divergence by maximizing 
$\mathcal{\Tilde{P}}_I(x_{i+1} | x_{1:i})$ 
to $1$ and minimizing $\mathcal{\Tilde{P}}(x_{i+1} | x_{1:i})$ to $\frac\sigma{d}$
so that
\begin{align*}
    D_{KL}(
        \mathcal{\Tilde{P}}_I(x_{i+1} | x_{1:i})
        \|
        \mathcal{\Tilde{P}}(x_{i+1} | x_{1:i})
    ) &= \sum_{x_{i+1}} \mathcal{\Tilde{P}}_I(x_{i+1} | x_{1:i})
    \log\frac{\mathcal{\Tilde{P}}_I(x_{i+1} | x_{1:i})}{\mathcal{\Tilde{P}}(x_{i+1} | x_{1:i})} \\
    &\leq \log\frac{d}{\sigma} 
\end{align*}

We can maximize  $|Z_i - Z_{i-1}|$ by maximizing the first term and minimizing the second term, or vice versa. In the first case, 
$|Z_i - Z_{i-1}|\leq |\log\frac{d}\sigma - \log\frac\sigma{d}|=2|\log\frac{d}{\sigma}|$. In the other case, $|Z_i - Z_{i-1}| \leq |0-\log\frac{d}\sigma| = |\log\frac{d}\sigma|$. 

Therefore, $|Z_i - Z_{i-1}| \leq c_i$ where $c_i = 2|\log\frac{d}{\sigma}|$. \qed

\begin{lemma}
\label{lemma:azuma}
By Azuma's inequality, with probability $1-\delta$, we have that $\|Z_T\| \leq b$
where $b=2 \log\frac{d}{\sigma} \sqrt{-8T  \log \frac{1}{\delta}}$
\end{lemma}
\proof
By Azuma's inequality, for all positive reals $b$,
\begin{align*}
    P(Z_T-Z_1\geq b) &\leq \exp\left(\frac{-b^2}{2\sum_{k=2}^T c^2_k}\right) \\
    P(Z_T-Z_1\leq b) &\geq 1 - \exp\left(\frac{-b^2}{2\sum_{k=2}^T c^2_k}\right)\\
     &\geq 1 - \exp\left(\frac{-b^2}{8\sum_{k=2}^T \log^2\frac{d}{\sigma}}\right)
\end{align*}

We can rewrite in terms of $\delta =  \exp\left(\frac{-b^2}{8\sum_{k=2}^T \log^2\frac{d}{\sigma}}\right)$ so
\begin{align*}
    b &= \sqrt{-\left({8\sum_{k=2}^T \log^2\frac{d}{\sigma}}\right) \log \delta}  \\
    &\leq \log\frac{d}{\sigma} \sqrt{-8T  \log \frac{1}{\delta}}  
\end{align*}

Therefore, $$P(Z_T-Z_1\leq b) \geq 1 - \delta$$ \qed

\section{Model Architecture Details}

In addition, we add a learnable scaling and bias parameter to the result of the embedding layer, so that the model can  still learn to scale it as needed.

\section{Convergence on other datasets}

\begin{figure}[tbh]
\centering
    \includegraphics[width=0.8\linewidth]{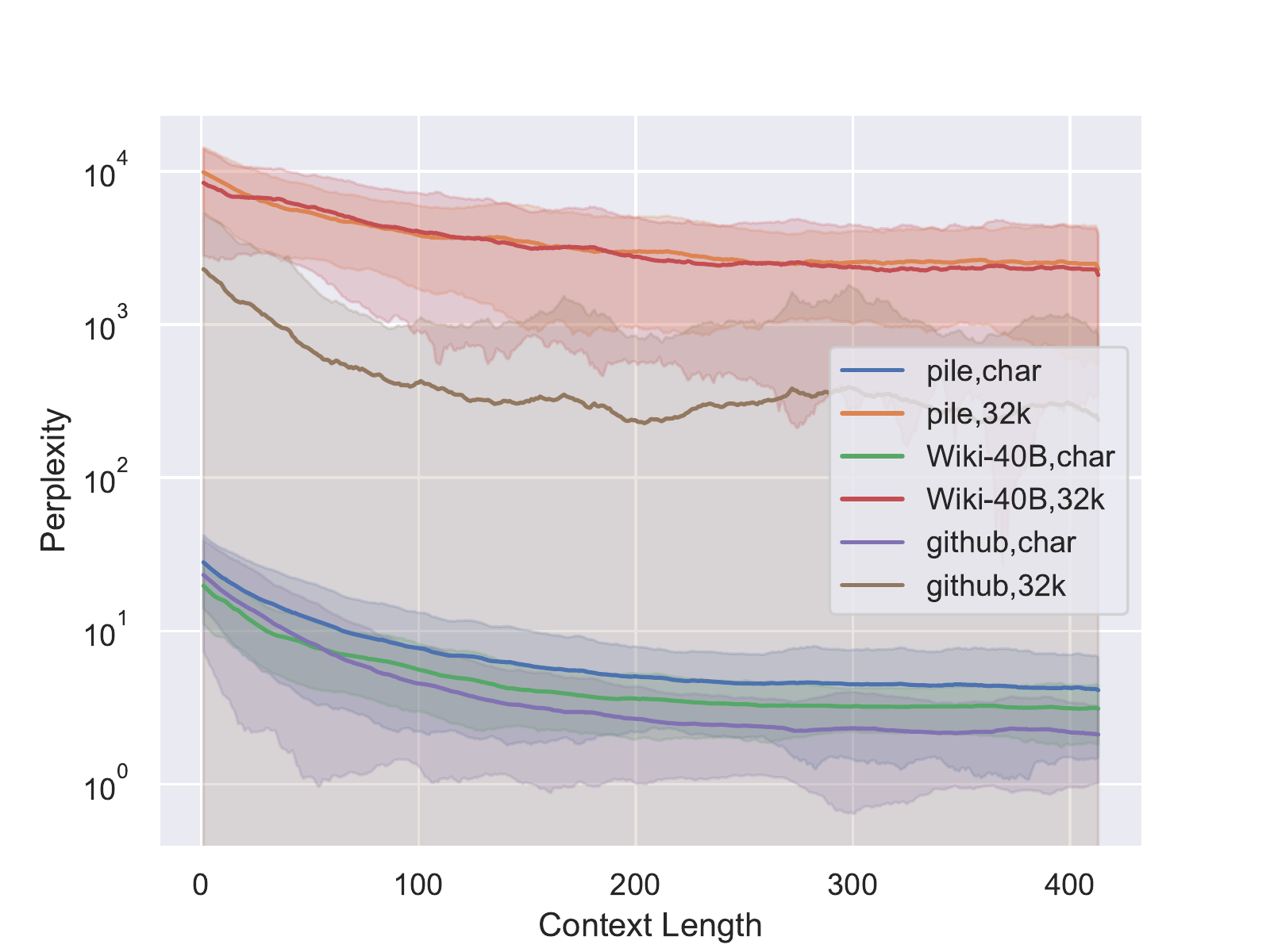}
    \caption{Smoothed Token Perplexity over the Pile, Wiki-40B and Github, with character-level and T5 default vocab}
    \label{fig:convegence3}
\end{figure}

Figure \ref{fig:convegence3} shows the perplexity of \modelname LMs across the three different datasets.  Note that Github converges significantly faster than standard Engish text like Wiki-40B, since code is more structured and easier to decipher the token permutation.

\section{Code Deciphering Full Examples}

Java: 
\begin{lstlisting}
binary_search()z
  if (high >= low)z
    mid = (high + low) / 2;
    if (arr[mid] == x)
      return mid;
    if (arr[mid] > x)z
      high = mid - 1;
      return binary_search();
    } elsez
      low = mid + 1;
      return binary_search();
    }
  } elsez
    return -1;
  }
}
void func2()z
\end{lstlisting}

Python: 
\begin{lstlisting}
binary_search()z
  if (high >= low)z
    mid = (high + low) // 2
    if (arr[mid] == x)z
      return mid
    if (arr[mid] > x)z
      high = mid - 1 
      return binary_search() 
    elsez
      low = mid + 1
      return binary_search()
  elsez
      return -1
def func2()z
\end{lstlisting}
\section{Semantic Deciphering Full Example}
\texttt{'crash!' 'aaah!' i looked up from my cup of coffee. 'crash!' - that was the cafe window. and 'aaah!' - that was kate. people in the cafe shouted. kate and i ran to the window. there was no one there. then i turned to kate and put my arm around her. 'are you all right?' i asked. 'yes,' she said. 'i think so.'
'what is it?' some one shouted and a short red-faced man ran into the room. the man took my arm. 'matt! what are you doing to kate?' he asked. 'nothing, papa,' kate replied. 'it wasn't him. it was from out in the street.' the red-faced man looked at the window and then at me. he turned to his daughter. 'are you ok, kate?' he asked. kate gave him a little smile. 'yes, i think i am, papa,' she said. then her father spoke to me. 'sorry, matt. i heard kate and i thought...' 'that's ok, paolo,' i answered.
it was ok. you see, this is soho, in the centre of london. in the day it's famous for music and films. at night people come and eat and drink in the restaurants. expensive restaurants and cheap restaurants; italian restaurants and chinese restaurants. and day and night there are internet cafes like the web cafe.
in soho you can buy any thing and any one. there are lots of nice people in soho. but there are also lots of people who are not very nice. i know because i live and work here. i often take a drink to a shop or cafe. i'm not rich and famous. and i don't know a lot. but i do know soho.
what one here is a drink
- restaurants
- music
- coffee
- father
the one here that drink is}

Example prediction of the \modelname with 32k vocabulary train on the Pile: 

\texttt{ - coffee. and i}

\section{Synthetic Reasoning Task}

Table~\ref{tab:syn-freq} shows a variant of the synthetic reasoning task results in Subsection~\ref{tab:syn}, where the symbols are instead sampled proportion to the token frequencies. Although 
the improvement still generally holds, the standard LM with character-based vocabulary  becomes significantly better. We believe that this is because the model can get a significant advantage by guessing among the most common letter. 

\begin{table}[htb]
    \centering
    \begin{tabular}{ll cccc}
    \toprule
          Dataset & Vocab & \multicolumn{2}{c}{LookUp Acc} & \multicolumn{2}{c}{Permutation Acc}\\
         \cmidrule(lr){3-4} \cmidrule(lr){5-6}
          & & Standard & LI & Standard & LI \\
          \midrule
          \multirow{2}{*}{Pile} &char & 72.80 & 90.95 & 40.63 & 60.47 \\
         &32k & 61.20 & 90.95 & 40.55 & 54.55 \\
         \midrule
        \multirow{2}{*}{Wiki-40B} &char & 75.55 & 63.45 & 42.71 & 59.86 \\
        &32k & 41.05 & 57.95 & 26.81 & 51.86 \\
        \midrule
         \multirow{2}{*}{Github} &char & 66.00 & 86.75 & 36.62 & 70.77 \\
          &32k & 59.25 & 78.45 & 37.46 & 65.04 \\
          \bottomrule
    \end{tabular}
    \caption{Synthetic Reasoning Tasks (adjusted for token frequencies)}
    \label{tab:syn-freq}
\end{table}

\section{ Language Models Regularized with Lexinvariance and BIG-bench Results}\label{bigbench}

As described in the main paper, we implement a  lexinvariance regularized Model in a way similar to embedding dropout.
Note that one problem in implementing it naively by using random Gaussian embeddings and learned embedding in a mixture is that the two would become quickly distinguishable from each other during training since learned embeddings often have larger norms, allowing the model simply ignore the randomized tokens. 
So instead of using random Gaussian embedding matrices in place of a learned embedding matrix, we explored another approach for training a \modelname regularized LM: training a standard LM with learnable embedding matrix over sequences partially applied with a random token permutation $B_p(x_1, \pi), \dots, B_p(x_1, \pi)$, where $B_p(x_i, \pi) = \pi(x_i)$ with probability $p$ and $B_p(x_i, \pi) = x_i$ with probability $1-p$. Since each token can be remapped to any other token with equal chance, the produced model should ideally also be \modelname when $p = 1$, though with no strict guarantees. In practice, we found the models trained this way behave very similarly to models with random Gaussian embedding.

We evaluate our model over BIG-bench tasks where the language model performance scales well, and we prioritize evaluating generative tasks over multiple-choice tasks. Tasks we evaluated on:

gre reading comprehension.mul, linguistics puzzles.gen, linguistics puzzles.gen, rhyming.gen, tellmewhy.gen, simple arithmetic multiple targets json.gen, simple arithmetic json subtasks.gen, disfl qa.gen, arithmetic.gen, bridging anaphora resolution barqa.gen, matrixshapes.gen, sufficient information.gen, logical args.mul, novel concepts.mul, code line description.mul, unnatural in context learning.gen, unit interpretation.mul, english proverbs.mul, general knowledge.mul, geometric shapes.gen, human organs senses.mul, contextual parametric knowledge conflicts.gen, crass ai.mul, auto categorization.gen, penguins in a table.gen, hindu knowledge.mul, english russian proverbs.mul, modified arithmetic.gen, cryobiology spanish.mul, evaluating information essentiality.mul, intent recognition.mul, understanding fables.mul, figure of speech detection.mul, empirical judgments.mul, simple ethical questions.mul, swahili english proverbs.mul, language identification.mul, phrase relatedness.mul, nonsense words grammar.mul, undo permutation.mul, object counting.gen, identify odd metaphor.mul, elementary math qa.mul, social iqa.mul, parsinlu qa.mul, metaphor understanding.mul, timedial.mul, causal judgment.mul, list functions.gen, implicatures.mul, date understanding.mul, codenames.gen, fact checker.mul, physics.mul, abstract narrative understanding.mul, emojis emotion prediction.mul, metaphor boolean.mul, strategyqa.gen, ascii word recognition.gen, auto debugging.gen, cause and effect.mul, conlang translation.gen, cryptonite.gen, cs algorithms.mul, dyck languages.mul, gender inclusive sentences german.gen, hindi question answering.gen, international phonetic alphabet transliterate.gen, irony identification.mul, logical fallacy detection.mul, movie dialog same or different.mul, operators.gen, paragraph segmentation.gen, parsinlu reading comprehension.gen, repeat copy logic.gen, rephrase.gen, simple arithmetic json.gen, simple arithmetic multiple targets json.gen, sports understanding.mul, word unscrambling.gen, hyperbaton.mul, linguistic mappings.gen, anachronisms.mul, indic cause and effect.mul, question selection.mul, hinglish toxicity.mul, snarks.mul, vitaminc fact verification.mul, international phonetic alphabet nli.mul, logic grid puzzle.mul, natural instructions.gen, entailed polarity.mul, list functions.gen, conceptual combinations.mul, goal step wikihow.mul, logical deduction.mul, conlang translation.gen, strange stories.mul, odd one out.mul, mult data wrangling.gen, temporal sequences.mul, analytic entailment.mul, disambiguation qa.mul, sentence ambiguity.mul, swedish to german proverbs.mul, logical sequence.mul, chess state tracking.gen, reasoning about colored objects.mul, implicit relations.mul, riddle sense.mul, physical intuition.mul, simple arithmetic json multiple choice.mul, geometric shapes.gen, gem.gen, simp turing concept.gen, common morpheme.mul, qa wikidata.gen, international phonetic alphabet transliterate.gen, similarities abstraction.gen, rephrase.gen, emoji movie.gen, qa wikidata.gen, word sorting.gen, emoji movie.gen, qa wikidata.gen, periodic elements.gen, hindi question answering.gen

Bellow, we plot the net percentage of tasks improved/deproved in each of the BIG-bench categories, out of the tasks that are changed by at least a threshold amount.

\begin{figure}
    \centering
    \includegraphics[width = \linewidth]{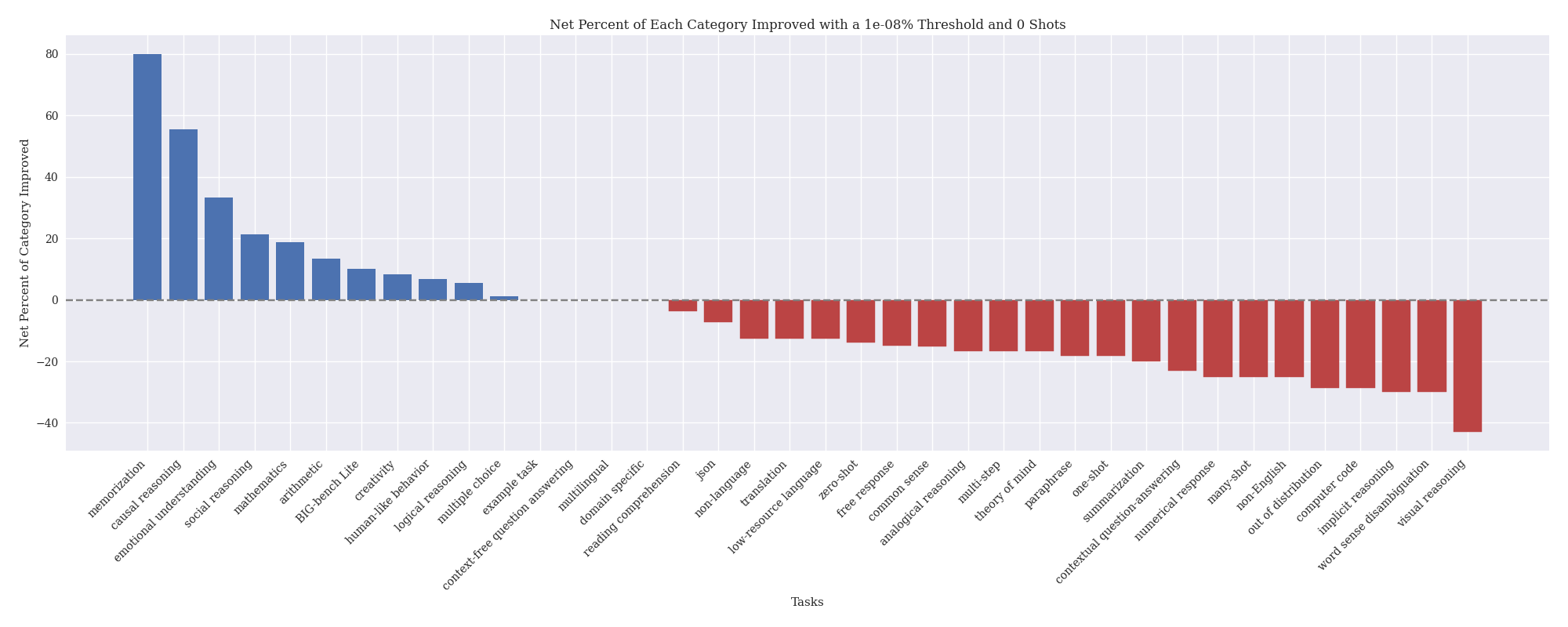}
    \includegraphics[width = \linewidth]{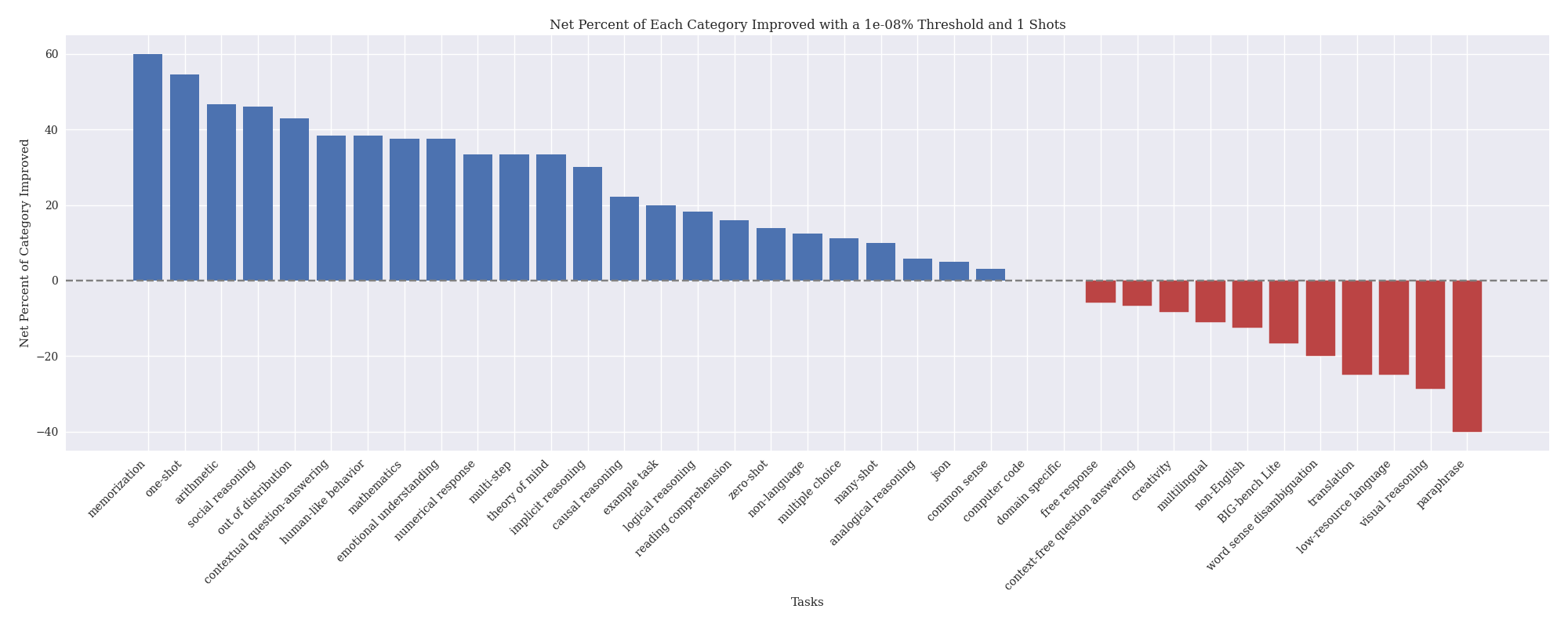}
    \includegraphics[width = \linewidth]{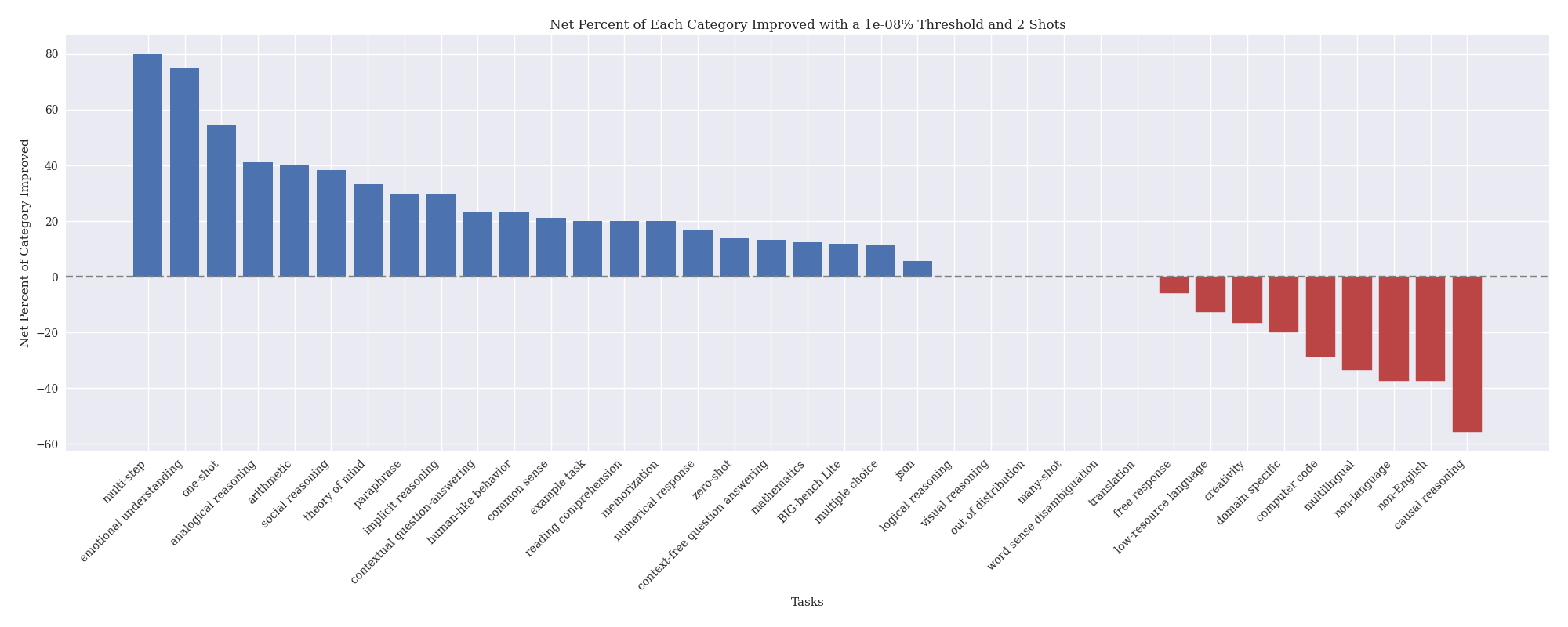}
    \includegraphics[width = \linewidth]{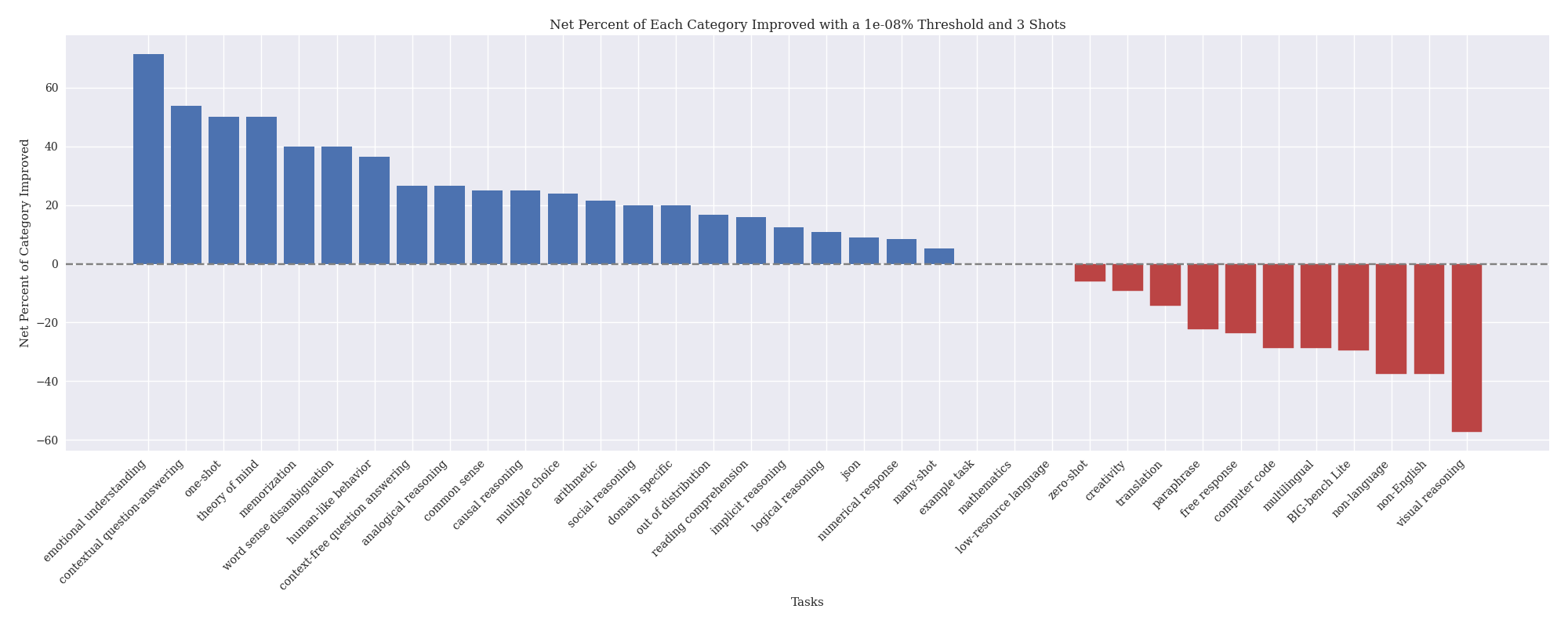}
    \caption{BIG-bench results with 0,1,2 and 3 shots.}
    \label{fig:my_label}
\end{figure}

\section{Compute}

We use one TPU v3-8 for all our pretraining runs. It takes approximately 23 hours for each pretraining run.

\section{Broader Impacts}
Our work primarily provides a scientific exploration and understanding of the properties of \modelname language models. More broadly, these properties could potentially help improve the robustness, generalizability, and reasoning ability of LMs in the future works. In general we don't foresee more specific negative societal impacts from this work other than general misuse of language models.

\end{document}